\documentclass[10pt,journal,compsoc]{IEEEtran}

\usepackage{cite}
\usepackage{amsmath,amssymb,amsfonts}
\usepackage{algorithmic}
\usepackage{graphicx}
\usepackage{textcomp}
\usepackage{xcolor}
\usepackage{diagbox} 
\usepackage{graphicx}
\usepackage{subfigure} 
\usepackage{hyperref}
\usepackage{caption2}

\newcommand{\fw}[1]{{\color{blue}{#1}}}

\begin{document}

\title{Interactive Reinforcement Learning for Feature Selection with Decision Tree in the Loop}

\author {Wei Fan, Kunpeng Liu, Hao Liu, Yong Ge, Hui Xiong~\IEEEmembership{Fellow,~IEEE}, and Yanjie Fu
\IEEEcompsocitemizethanks{
\IEEEcompsocthanksitem Wei Fan, Kunpeng Liu and Yanjie Fu are with the Department of Computer Science, University of Central Florida. \protect\\
E-mail: \{weifan, kunpengliu\}@knights.ucf.edu, yanjie.fu@ucf.edu
\IEEEcompsocthanksitem Hao Liu is with Business Intelligence Lab, Baidu Research. \protect\\
E-mail: liuhao30@baidu.com
\IEEEcompsocthanksitem Yong Ge is with the Eller College of Management, University of Arizona. \protect\\
E-mail: yongge@arizona.edu
\IEEEcompsocthanksitem Hui Xiong is with Rutgers University.
E-mail: hxiong@rutgers.edu }
}

\IEEEtitleabstractindextext{%
\begin{abstract}
We study the problem of balancing effectiveness and efficiency in automated feature selection. Feature selection is to find an optimal feature subset from large feature space. 
After exploring many feature selection methods, we observe a computational dilemma: 1) traditional feature selection (e.g., mRMR) is mostly efficient, but difficult to identify the best subset; 2) the emerging reinforced feature selection automatically navigates feature space to search the best subset, but is usually inefficient. Are automation and efficiency always apart from each other? 
Can we bridge the gap between effectiveness and efficiency under automation?
Motivated by this dilemma, we aim to develop a novel feature space navigation method. In our preliminary work, we leveraged interactive reinforcement learning to accelerate feature selection by external trainer-agent interaction. Our preliminary work can be significantly improved by modeling the structured knowledge of its downstream task (e.g., decision tree) as learning feedback.
In this journal version, we propose a novel interactive and closed-loop architecture to simultaneously model interactive reinforcement learning (IRL) and decision tree feedback (DTF). Specifically, IRL is to create an interactive feature selection loop and DTF is to feed structured feature knowledge back to the loop. The DTF improves IRL from two aspects. First, the tree-structured feature hierarchy generated by decision tree is leveraged to improve state representation. In particular, we represent the selected feature subset as an undirected graph of feature-feature correlations and a directed tree of decision features. We propose a new embedding method capable of empowering Graph Convolutional Network (GCN) to jointly learn state representation from both the graph and the tree. Second, the tree-structured feature hierarchy is exploited to develop a new reward scheme. In particular, we personalize reward assignment of agents based on decision tree feature importance. In addition, observing agents' actions can also be a feedback, we devise another new reward scheme, to weigh and assign reward based on the selected frequency ratio of each agent in historical action records.
Finally, we present extensive experiments with real-world datasets to demonstrate the improved performances of our method.


\end{abstract}

}

\maketitle

\IEEEdisplaynontitleabstractindextext

%
\IEEEpeerreviewmaketitle

\ifCLASSOPTIONcompsoc
\IEEEraisesectionheading{\section{Introduction}\label{sec:introduction}}
\else
\vspace{-0.3cm}
\section{Introduction}
\vspace{-0.1cm}
\label{sec:introduction}
\fi
We aim to study the problem of balancing effectiveness and efficiency in automated feature selection. 
Feature selection is to find the optimal feature subset from large feature space, which is essential for 
lots of machine learning tasks. 


Classic feature selection methods include: filter methods (e.g., univariate selection \cite{forman2003extensive}, correlation based  selection \cite{hall1999feature}), wrapper methods (e.g., branch and bound algorithms \cite{narendra1977branch}), and  embedded methods (e.g., LASSO \cite{tibshirani1996regression }). 
{ Recently, the emerging reinforced feature selection methods~\cite{fard2013using,kroon2009automatic,liu2019automating}} formulate feature selection into a Reinforcement Learning (RL) task, in order to automate the selection process.
Our preliminary study~\cite{fan2020autofs} has observed a computational dilemma in feature selection:
(1)  classic selection methods are mostly efficient, but difficult to identify the best subset; 
2) the emerging reinforced selection methods automatically navigate feature space to explore the best subset, but are usually inefficient. 
Are automation and efficiency always apart from each other? Can we strive for a balance between effectiveness and efficiency under automation?

\begin{figure}[th]
\centering
\centering
\vspace{-2mm}
\includegraphics[width=7.8cm,height=3.8cm]{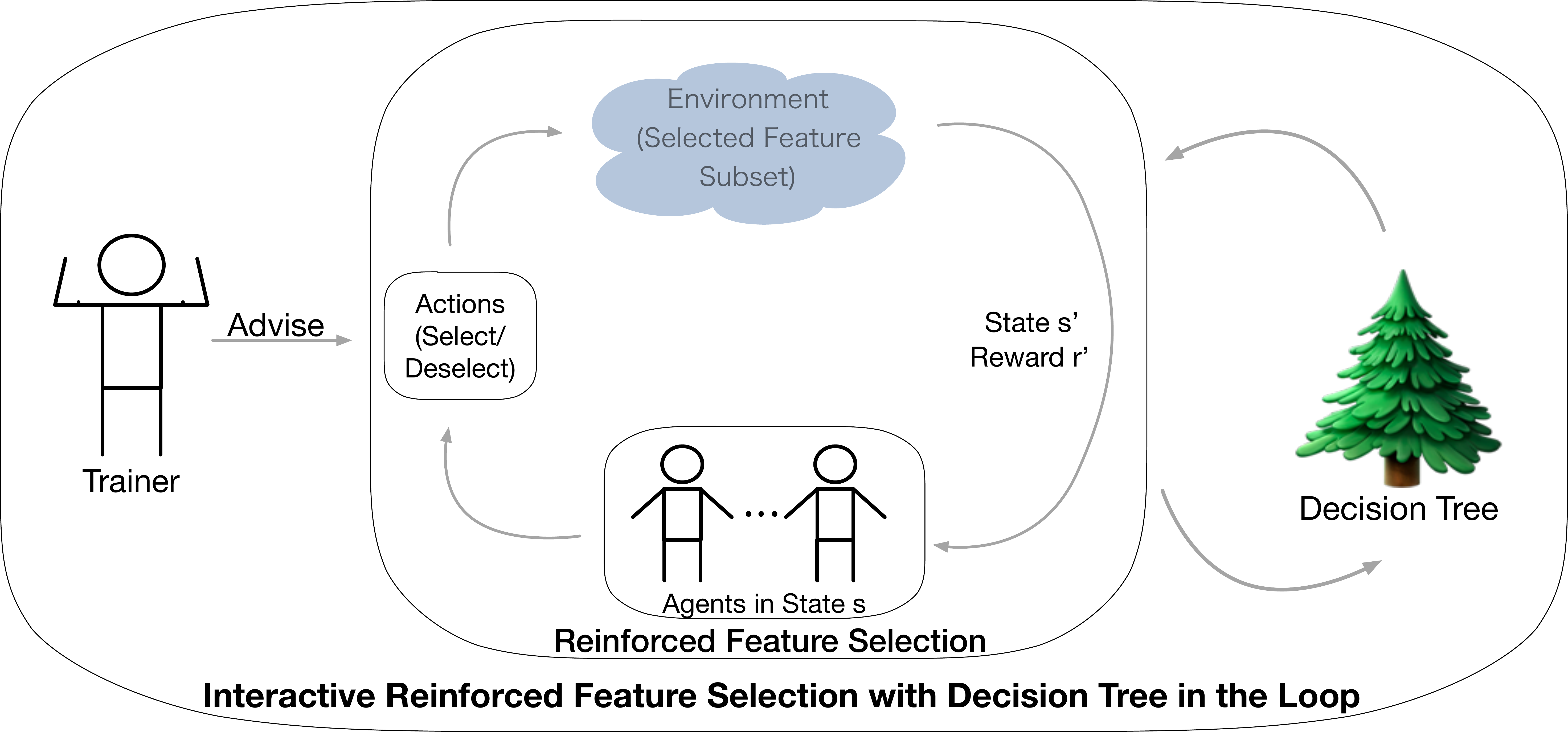}
\vspace{-2.8mm}
\caption{
Interactive reinforced feature selection via interaction with external trainers and downstream tasks (decision tree).
}
\label{intro1}
\vspace{-4.5mm}
\end{figure}

Motivated by the above dilemma, our preliminary work~\cite{fan2020autofs}  integrates self-exploration experience of regular RL and external skilled trainers via interaction to address the problem. 
We developed a diversity-aware interactive reinforcement learning (IRL) approach. 
In this approach, we formulated feature selection as a multi-agent reinforcement learning task, where an agent is a feature selector that selects or deselects its corresponding feature. 
We integrated the interactive learning feature into multi-agent RL, in order to introduce prior knowledge of external trainers. 
We identified an interesting property of interaction: diversity matters.

As a result, we diversified the set of external trainers by adopting two traditional feature selection methods:  KBest and Decision Tree as external trainers.
We diversified advice for agents by dynamically selecting a group of agents to participate in feature selection, which we call participated agents.
We ensured different participated agents to receive personalized and tailored advice. In particular, we categorized the participated agents into: assertive and hesitant agents (features). The assertive agents are more confident about their selection decisions, and  thus don't need advice from trainers. The hesitant agents are less confident about their decisions, and  thus need advice from trainers. 
Also, to diversify the teaching process, we propose a Hybrid Teaching strategy, to iteratively let various trainers take the teacher role at different stages (e.g., early, middle, late). Such strategy can fuse the experience of the trainers and thus provide better advice.

The preliminary study~\cite{fan2020autofs} addressed the interaction between external trainers and agents. This paper will address another important research question: Can the interaction between downstream predictive tasks and RL, further improve feature selection? How does a downstream predictive task, for example, a decision tree, provide feature structured knowledge to create a feedback-improvement loop?

To fill the gap, we develop a framework (Figure \ref{intro1}) that unifies both interactions between external trainers and agents, and interaction between downstream tasks and RL into a joint and interactive architecture. 
Indeed, many downstream predictive tasks, such as decision tree, random forest, LASSO, not just output predictive targets, but also produce structured knowledge of features. 
We propose to feed such structured feature knowledge back to reinforced feature selection. 
Taking decision tree as an example downstream task. 
The feedback-improvement loop of decision tree will improve the effectiveness of reinforced feature selection from three perspectives. 
First, in our preliminary study, we analogize a feature as a node to create a feature-feature similarity graph, and then exploit the Graph Convolutional Network (GCN) model to learn the graph embedding as the state representation of environment. 
While the GCN over feature-feature graph has achieved relatively good performances, we are particularly interested in: when we use the decision tree as the downstream task, what role does the tree-structured feedback play in the feature-feature graph based GCN? 
The tree-structured feature importance hierarchy of the decision tree  reflects the latent feature correlations in the selected feature subspace. 
Based on this insight, we propose to jointly learn the state representation of the feature subset (environment) from not just  an undirected graph of feature-feature correlations, but also a directed tree graph of decision features. 
Second, we observe that a decision feature importance tree is a subgraph of the feature-feature graph. 
We develop an improved version of graph convolutional network (GCN) to empower GCN to pay specific attention to the tree when preserving the structure information of the graph.
Third, unlike equally sharing of reward, we devise a new personalized reward scheme to better measure agent reward assignment, based on the feature importance from the decision tree. 
In particular, 
the feature importance feedback from decision tree describes how good the action is for an agent to select a specific feature. Thus, an action that selects more important feature should receive higher reward.
In addition, we propose another reward scheme based on the historical action records. In particular, this scheme weighs and assigns reward based on the selected frequency ratio of each feature. Generally, agents are more likely to choose advantageous actions for long-term reward maximization; as a result, features' selected frequency ratio is another feedback to reflect feature importance and thus can be used to assign reward.


In summary, in this paper, we develop a joint and interactive  architecture for reinforced feature selection. 
Specifically, our contributions are as follows: 
1) We formulate the  feature selection problem into an interactive reinforcement learning framework. 
2) We develop a joint and interactive architecture to unify both interaction between feature agents and external trainers, and the  interaction between downstream task and reinforcement learning
3) To model the interaction between agents and external trainers, we devise a new diversity-aware mechanism including
(a) diversifying the external trainers; 
(b) diversifying the advice that agents will receive; 
(c) diversifying the set of agents that will receive advice; 
(d) diversifying the teaching process.
4) To model the interaction between downstream task and reinforcement learning, we develop an improved GCN to jointly learn precise state representation from both feature-feature similarity graph and feature importance tree, and design two new reward schemes that personalize assigned rewards to multiple agents. 
5) We conduct extensive experiments on real-world datasets, and the results demonstrate the advantage of our methods.

\fw{
}


\vspace{-0.3cm}
\section{Preliminary}
\vspace{-0.1cm}

\begin{table}
\vspace{-0mm}
\scriptsize
\centering
\caption{Notations.}
\vspace{+0.5mm}
\begin{tabular}{p{0.08\textwidth}p{0.34\textwidth}}
\hline
Notations & Definition \\
\hline
$|\cdot|$ &  The cardinality of a set \\
$\lceil x \rceil$ &    The greatest integer less than or equal to x \\
$\overline{x}$ &   The complement of set x   \\
$N$ &  The number of features \\
$f_i$ &  The $i$-th feature \\
$agt_i$ &  The $i$-th agent \\
$\pi_i$ & The policy network of the $i$-th agent\\
$a_i$ & The action performed by the $i$-th agent \\
$r_i$ &  The reward assigned to the $i$-th agent \\
\hline
\end{tabular}
\label{method0}
\vspace{-5mm}
\end{table}

We introduce definitions and problem statement of reinforced feature selection when Interactive Reinforcement Learning (IRL) is applied.
Then, 
we show the overview of our framework with decision tree in the loop. Table \ref{method0} shows some commonly used notations.

\begin{figure*}[h]
\centering
\includegraphics[width=16.2cm,height=7.0cm]{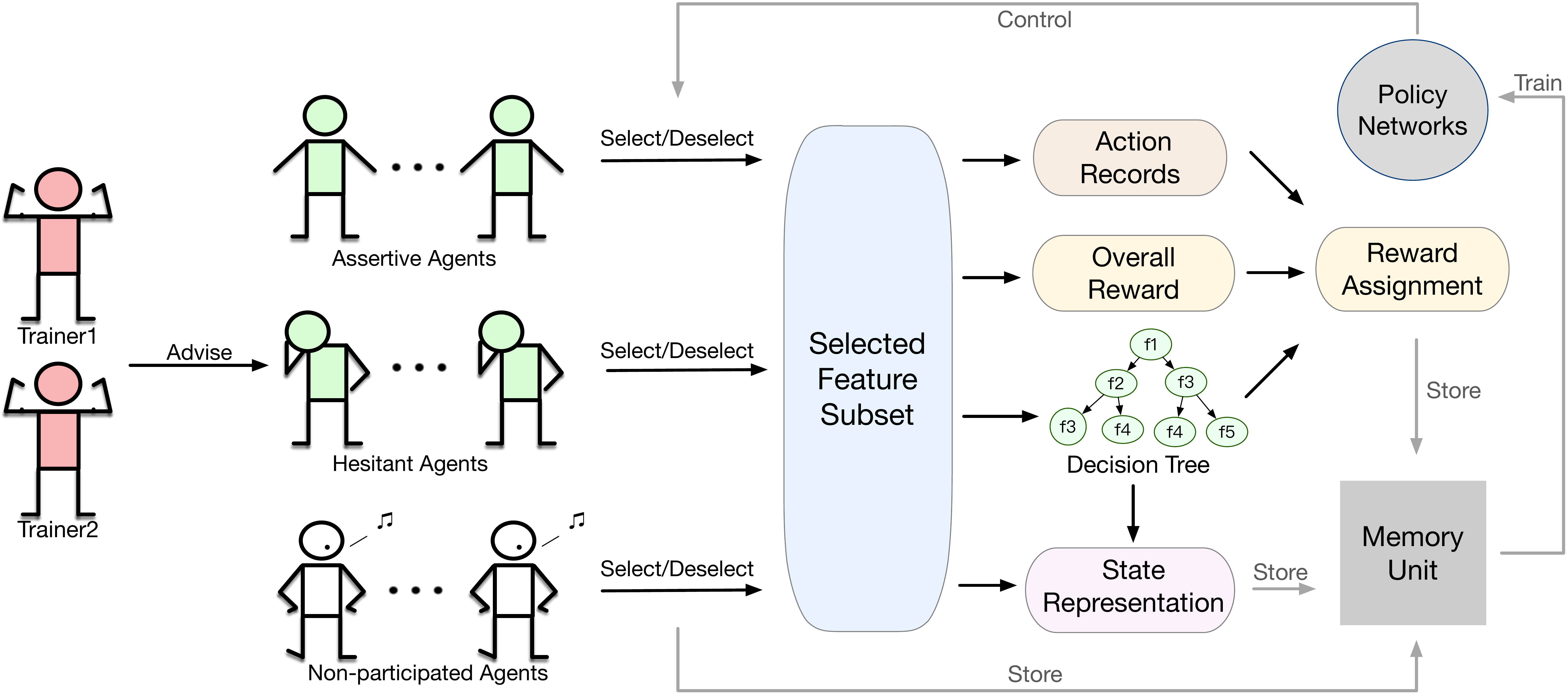}
\caption{Framework Overview. Agents select/deselect their corresponding features based on their policy networks and trainers' advice. Action, reward and state are stored in memory unit to train policy networks. The structure feedback of decision tree feeds to improve state representation. Decision tree hierarchy and action records help reward measurement.}
\label{overview}
\vspace{-2.5mm}
\end{figure*}

\subsection{Definitions and Problem Statement}

\textit{Definition 2.1.} \textbf{Agents.} Each feature $f_i$ is associated with an agent $agt_i$. After observing a state of the environment, agents use their policy networks to make decisions on the selection of their corresponding features.

\noindent\textit{Definition 2.2.} \textbf{Actions}. Multiple agents corporately make decisions to select a feature subset. For a single agent, its action space $a_i$ contains two actions, i.e., select and deselect.



\noindent\textit{Definition 2.3.} \textbf{State.} The state $s$ is defined as the representation of the environment, which is the selected feature subset. 
More details of state representation are in Section \ref{sec_state_representation}.

\noindent\textit{Definition 2.4.} \textbf{Reward.} The reward is to inspire the feature subspace exploration process. We firstly derive the overall reward from 
the selective feature subset. Then we assign the overall reward to multiple agents using different strategies. More details of reward scheme are in Section \ref{method_reward_part}.




\noindent\textit{Definition 2.5.} \textbf{Trainer.} In the apprenticeship of reinforcement learning, the actions of agents are immature, and thus it is important to give some advice to the agents. We define the source of the advice as trainers.

\noindent\textit{Definition 2.6.} \textbf{Advice.}  Trainers give multiple agents advice on their actions;
gents will follow the advice to take actions.

\noindent\textit{Definition 2.7.} \textbf{Problem Statement.} In this paper, we study the feature selection problem with interactive reinforcement learning. Formally, given a set of features $F =  \{f_1, f_2,  ..., f_N \}$ where $N$ is the number of features, our aim is to find an optimal feature subset $F' \subseteq F $ which is most appropriate for the downstream task. 

In this paper, considering the existence of $N$ features, we create $N$ agents $ \{ agt_1, agt_2, ..., agt_N \} $ correspondingly for feature $\{f_1, f_2,  ..., f_N \}$. Each agent uses its own policy network $\{ \pi_1, \pi_2, ..., \pi_N \}$ to make decisions to select or deselect its corresponding feature, where the actions are denoted by $\{ a_1, a_2, ..., a_N   \}$. For $i \in [1,N],  a_i = 1$ means agent $agt_i$ decides to select feature $f_i$; $a_i = 0$ means agent $agt_i$ decides to deselect feature $f_i$. Whenever actions are issued in a step, the selected feature subset changes, and then we can derive the changed state $s$. Finally, the reward $\{r_1, r_2, ..., r_N\}$ is assigned to the agents based on their actions.

\vspace{-0mm}
\subsection{Framework Overview}
Before the framework illustration, we first introduce some basic components of our framework.

\noindent{\textbf{1) Participated/Non-participated Features (Agents).}} We propose to use classical feature selection methods as trainers. However, 
the trainers only provide similar or the same advice to agents every time.
To solve the problem and diversify the advice, we dynamically change the input to the trainer by selecting a set of features, which we call participated features. We define the participated features as those features that were selected by agents in last step, e.g., if at step $t-1$ agents select $f_2, f_3,f_5$, the participated features at step $t$ are $f_2, f_3, f_5$. The corresponding agents of participated features are participated agents; other agents are non-participated agents which select/deselect their corresponding non-participated features.

\noindent{\textbf{2) Assertive/Hesitant Features (Agents)}}. We dynamically divide the participated features into \textbf{assertive features} and \textbf{hesitant features}, whose corresponding  agents  are  accordingly  called \textbf{assertive agents} and \textbf{hesitant agents}. Specifically, at step $t$, the participated features are divided into: assertive features, defined as the features decided to be selected by the policies, and hesitant features, defined as the features decided to be deselected by the policies. For example, at step $t$ participated features are $f_2,f_3,f_5$, and policy networks $\pi_2,\pi_3,\pi_5$ decide to select $f_2$ and deselect $f_3, f_5$. Then, assertive features are $f_2$; assertive agents are $agt_2$; hesitant features are $f_3,f_5$; hesitant agents are $agt_3,agt_5$.

\noindent{\textbf{3) Initial/Advised Actions}}. In each step, agents use their policy networks to ﬁrstly make action decisions, which we call \textbf{initial actions}. Then, agents take advice from external trainers and update their actions; the actions advised by trainers are called \textbf{advised actions}. In this framework, only hesitant agents need to follow the advice and take the advised actions. 

\noindent{\textbf{4) Decision Tree}}.  In the automated feature selection, the downstream task is to evaluate whether the selected features are good or not. When we use decision tree as the downstream task, the rich structure information of the tree can provide much feedback to the upstream feature selection process. Thus, we make full use of the tree structure feedback to improve the reinforced feature selection.

\begin{figure}[h]
\centering
\includegraphics[width=8.5cm,height=8.5cm]{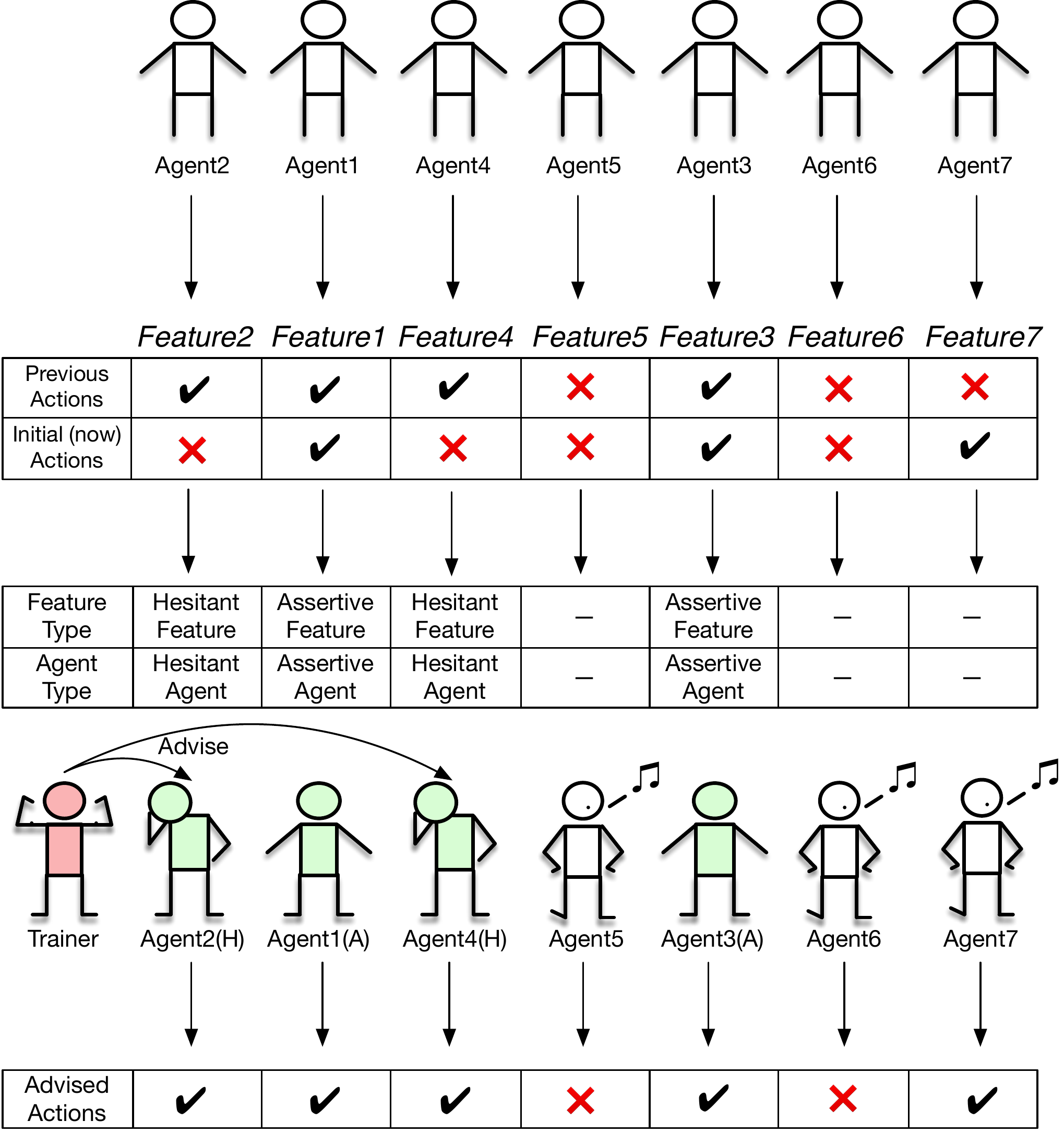}
\caption{General process of trainer guiding agents. Our framework firstly identifies assertive agents (labeled by `A') and hesitant agents (labeled by `H'). Then, the trainer offers advice to hesitant agents.}
\label{method1}
\vspace{-2mm}
\end{figure}

Figure \ref{overview} shows an overview of our framework.  There are multiple agents, each of which has its own Deep Q-Network (DQN) \cite{mnih2015human,mnih2013playing} as policy. At the beginning of each step,  each agent makes an initial action decision, from which we can divide all the agents into assertive agents, hesitant agents and non-participated agents. 
Then, the trainers come to provide action advice, and the hesitant agents change their initial actions to take advised actions.
After agents take actions, we derive a selected feature subset, whose representation is the state. We can train a decision tree on the selected features and use its structure feedback to improve the state representation.
Also, both the feature importance of the tree and the records of taken actions can be used to personalize the assigned reward to each agent, which is more reasonable than the traditional equally sharing of reward \cite{liu2019automating}. Then, a tuple of four components is stored into the memory unit, including the last state, the current state, the actions and the reward.
In the training process, for agent $agt_i$ at step $t$, we select mini-batches from memory unit to train the policy networks, in order to maximize the long-term reward based on Bellman Equation \cite{sutton2018reinforcement}.

\vspace{-0.3cm}
\section{Method}
\vspace{-0.1cm}

We first present details of the design of our interactive reinforced feature selection (IRFS) framework. Then, we introduce our proposed state representation methods based on the decision tree structure feedback. 
Finally, we present the detailed design of two personalized reward schemes.

\subsection{Interactive Reinforced Feature Selection}
Our design of our interactive reinforced feature selection (IRFS) includes three parts: (1) IRFS with KBest based trainer; (2) IRFS with Decision Tree based Trainer; (3) IRFS with Hybrid Teaching strategy. 
We illustrate the details of such design along this line.

\subsubsection{{Interactive Reinforced Feature Selection with KBest Based Trainer}}
\label{3.1}
We propose to formulate the feature selection problem into an interactive reinforcement learning framework called interactive reinforced feature selection (IRFS). 
Figure \ref{method1} illustrates the general process of how agents are advised by the trainer. In this formulation, we propose an advanced trainer based on a filter feature selection method, namely \textbf{KBest based trainer}. In our framework, KBest based trainer advises hesitant agents by comparing hesitant features with assertive features. We show how the trainer gives advice step by step as follows:

\noindent\textit{\textbf{1) Identifying Assertive/Hesitant Agents}}.
Given the policy networks $\{ {\pi_1}^t, {\pi_2}^t, ..., {\pi_N}^t  \}$ of multiple agents at step $t$, each agent makes an initial decision to select or deselect its corresponding feature; thus, we get an \textbf{initial} action list at step $t$, denoted by $\{ {{a_1}^t}', {{a_2}^t}', ..., {{a_N}^t}'\}$. We record actions that agents have already taken at step $t-1$, denoted by $\{ a_1^{t-1}, a_2^{t-1}, ..., a_N^{t-1} \}$. 
Then, we can find participated features at step $t$ by $F_p = \{\,f_i\,|\,i\in [1,N], a_i^{t-1} = 1 \}$. 
Also, we can identify assertive features as well as assertive agents. Assertive features are $F_a = \{\,f_i\,|\,i\in [1,N], f_i \in F_p \,\&\, {a_i^{t}}' = 1 \}$; assertive agents are $H_a = \{\,agt_i\,|\,i\in [1,N], f_i \in F_p \,\&\, {a_i^{t}}' = 1 \}$. 
Similarly, we can identify hesitant features and hesitant agents. i.e., hesitant features are $F_h = \{\,f_i\,|\,i\in [1,N], f_i \in F_p \,\&\, {a_i^t}' = 0 \}$; hesitant agents are $H_h = \{\,agt_i\,|\,i\in [1,N], f_i \in F_p \,\&\, {a_i^{t}}' = 0 \}$.



\begin{table}
\small
\centering
\begin{tabular}{p{0.45\textwidth}}
\hline
Algorithm 1: IRFS with KBest based trainer \\
\hline
\textbf{Input}: number of features: $N$, set of features: $\{ f_1, f_2, ..., f_N \}$, agent actions taken at step (t-1): $\{ a_1^{t-1}, a_2^{t-1}, ..., a_N^{t-1} \}$, policy networks at step $t$: $\{ \pi_1^t, \pi_2^t, ..., \pi_N^t  \}$, state at step $t$: $s$, K-Best Algorithm: $SelectK(input\, features, input\, k)$ \\
\textbf{Output}: advised actions at step $t$: $\{ a_1^{t}, a_2^{t}, ..., a_N^{t} \}$\\
initialize participated feature set $F_p$, assertive feature set $F_a$, hesitant feature set $F_h$\\
1: \;\textbf{for} $i$ = 1 to $N$ \textbf{do} \\
2: \;\;\;\;\; ${a_i^t}' \gets $  the highest-valued action in ${\pi_i}^t(s)$\\
3: \;\;\;\;\; \textbf{if} $a_i^{t-1} = 1$ \textbf{do} \\
4: \;\;\;\;\;\;\;\; add $f_i$ into $F_p$\\
5: \;\;\;\;\; \textbf{if} ${a_i^t}' = 1 \; \& \; a_i^{t-1} = 1$ \textbf{do} \\
6: \;\;\;\;\;\;\;\; add $f_i$ into $F_a$\\
7: \;\;\;\;\; \textbf{elseif} ${a_i^t}' = 0 \; \& \; a_i^{t-1} = 1$ \textbf{do} \\
8: \;\;\;\;\;\;\;\; add $f_i$ into $F_h$\\
9: \;\;integer $m \gets |F_a|$, integer $n \gets |F_h|$ \\
10: integer $k = \lceil m/2 + n \rceil$ \\
11: $F_{kbest} \gets SelectK(F_p, k) $\\
12: \textbf{for} $i$ = 1 to $N$ \textbf{do} \\
13: \;\;\; \textbf{if} $f_i \in F_h \;\&\; f_i \in F_{kbest}$ \textbf{do} \\
14: \;\;\;\;\;\; $a_i^t \gets \overline{{a_i^t}'}$ \\
13: \;\;\; \textbf{else} \textbf{do} \\
14: \;\;\;\;\;\; $a_i^t \gets {a_i^t}'$ \\
15: \textbf{return} $\{a_1^t, a_2^t, ..., a_N^t\}$ \\
\hline
\end{tabular}
\vspace{-3mm}
\end{table}

\noindent\textit{\textbf{2) Acquiring Advice from KBest Based Trainer}}.
After identifying assertive/hesitant agents, we propose a KBest based trainer, which can advise hesitant agents to update their initial actions. 
Our perspective is: if the trainer thinks a hesitant feature is even better than half of the assertive features, its corresponding agent should change the action from deselection to selection. 

\textit{ \textbf{Step1}: (Warm-up)} We obtain the number of assertive features by $m = |F_a|$, and the number of hesitant features by $n = |F_h|$. 
We set the integer $ k = \lceil m/2 + n \rceil$. Then, we use K-Best algorithm to select top $k$ features in $F_p$. We denote these $k$ features by $F_{KBest}$. 

\textit{ \textbf{Step2}: (Advise)} The indices of agents which need to change actions are selected by $I_{advised} = \{ \, i \, |\, i \in [1,N], \, f_i \in F_h \,\, and \,\, f_i \in  F_{kbest} \}$. 
Finally, we can get the advised action that $agt_i$ will finally take at step $t$, denoted by 
  
\begin{equation}
\centering
a_i^t=\left\{
\begin{aligned}
\overline{ {a_i^t}'}, i \in I_{advised}  \\
{a_i^t}', i \notin I_{advised} \\
\end{aligned}
\right.
\end{equation}

\begin{table}
\small
\centering
\begin{tabular}{p{0.45\textwidth}}
\hline
Algorithm 2: IRFS with Decision Tree based trainer \\
\hline
\textbf{Input}: number of features: $N$, set of features: $\{ f_1, f_2, ..., f_N \}$, agent actions taken at step $(t-1)$: $\{ a_1^{t-1}, a_2^{t-1}, ..., a_N^{t-1} \}$,  policy networks at step $t$: $\{ \pi_1^t, \pi_2^t, ..., \pi_N^t  \}$, state at step $t$: $s$, Decision Tree Classifier: $DecisionTree(input\,features)$ \\
\textbf{Output}: advised actions at step $t$: $\{ a_1^{t}, a_2^{t}, ..., a_N^{t} \}$\\
initialize hesitant feature set $F_h$, assertive feature set $F_a$\\
1: \;\textbf{for} $i$ = 1 to $N$ \textbf{do} \\
2: \;\;\;\;\; ${a_i^t}' \gets $  the highest-valued action in $\pi_i^t(s)$\\

3: \;\;\;\;\; \textbf{if} $a_i^{t-1} = 1$ \textbf{do} \\

4: \;\;\;\;\;\;\;\; add $f_i$ into $F_p$\\

5: \;\;\;\;\; \textbf{if} ${a_i^t}' = 1 \; \& \; a_i^{t-1} = 1$ \textbf{do} \\
6: \;\;\;\;\;\;\;\; add $f_i$ into $F_a$\\

7: \;\;\;\;\; \textbf{elseif} ${a_i^t}' = 0 \; \& \; a_i^{t-1} = 1$ \textbf{do} \\
8: \;\;\;\;\;\;\;\; add $f_i$ into $F_h$\\
 
9: \;  $T \gets DecisionTree(F_p)$ \\
10:  $\{imp_{f_{p_1}}, ..., imp_{f_{p_t}}\} \gets T.feature\_importances$ \\
11: $IMP_a \gets \{\; imp_{f_{p_j}}\;|\; f_{p_j} \in F_a  \},\; g \gets Median(IMP_a)  $  \\
12: $IMP_h \gets \{\; imp_{f_{p_j}}\;|\; f_{p_j} \in F_h  \}$\\
13: \textbf{for} $i$ = 1 to $N$ \textbf{do} \\
14: \;\;\; \textbf{if} $f_i \in F_h \;\&\; imp_{f_i} > g $ \textbf{do} \\
15: \;\;\;\;\;\; $a_i^t \gets \overline{{a_i^t}'}$ \\
16: \;\;\; \textbf{else} \textbf{do} \\
17: \;\;\;\;\;\; $a_i^t \gets {a_i^t}'$ \\
18: \textbf{return} $\{a_1^t, a_2^t, ..., a_N^t\}$ \\
\hline
\end{tabular}
\vspace{-4mm}
\end{table}

\subsubsection{{Interactive Reinforced Feature Selection with Decision Tree Based Trainer} }

In our IRFS framework, we propose another trainer based on a wrapper feature selection method, namely \textbf{Decision Tree based trainer}. The Decision Tree based trainer is similar to the KBest based trainer, both of which use trainer's evaluation on participated features to advise hesitant agents. However, the evaluation criteria of these two trainers are different. Decision Tree based trainer utilizes feature importance of a decision tree to give advice. Specifically, if the feature importance of any hesitant feature is larger than half of assertive features, this hesitant feature is considered comparatively important and thus should be selected.


The pipeline of IRFS with Decision Tree based trainer guiding agents can also be demonstrated in a two-phase process: (1) Identifying Assertive/Hesitant Agents; (2) Acquiring Advice from Decision Tree Based Trainer. 
The first phase is the same as phase one in Section \ref{3.1}, and the second phase is detailed as follows:

\textit{\textbf{Step1}: (Warm-up)} We denote the participated feature set as $F_p = \{f_{p_1}, f_{p_2}, ..., f_{p_t} \}$. We train a decision tree on $F_p$, and then get the feature importance for each feature, denoted by $ \{ imp_{f_{p_1}}, imp_{f_{p_2}}, ..., imp_{f_{p_t}} \}$. For hesitant features $F_h$, their importance $IMP_h = \{imp_{f_{p_j}}\;|\, f_{p_j} \in F_h  \}$; for assertive features $F_a$, their importance $IMP_a = \{imp_{f_{p_j}}\;|\, f_{p_j} \in F_a  \}$. We denote the median of $IMP_a$ by $g$.

\textit{\textbf{Step2}: (Advise)} The indices of agents which need to change actions are selected by $I_{advised} = \{\, p_j \,|\, f_{p_j} \in F_h \,\, and \,\, imp_{f_{p_j}} > g  \}$. Finally, we can get the advised action that $agt_i$ will finally take at step $t$, denoted by 

\begin{figure}[h]
\centering
\includegraphics[width=8cm,height=7cm]{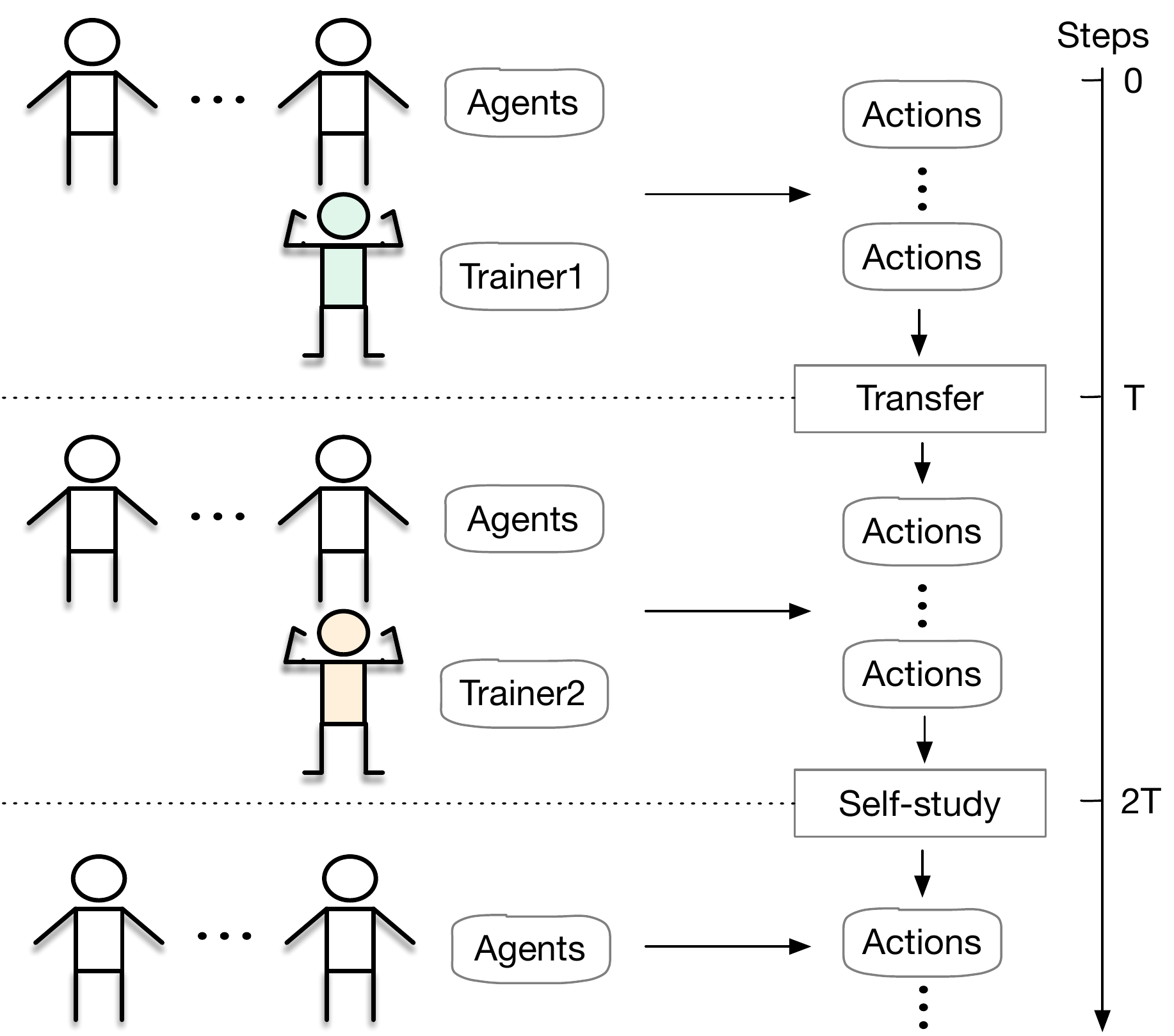}
\vspace{-1mm}
\caption{General process of Hybrid Teaching strategy. One trainer gives advice firstly; then, another trainer gives advice. Finally, agents explore and learn by themselves.}
\label{method2}
\vspace{-4mm}
\end{figure}

\begin{equation}
\centering
a_i^t=\left\{
\begin{aligned}
\overline{{a_i^t}'}, i \in I_{advised}  \\
{a_i^t}', i \notin I_{advised} \\
\end{aligned}
\right.
\end{equation}

\subsubsection{{Interactive Reinforced Feature Selection with Hybrid Teaching Strategy}}

\begin{figure}[h]
\centering
\includegraphics[width=8.5cm]{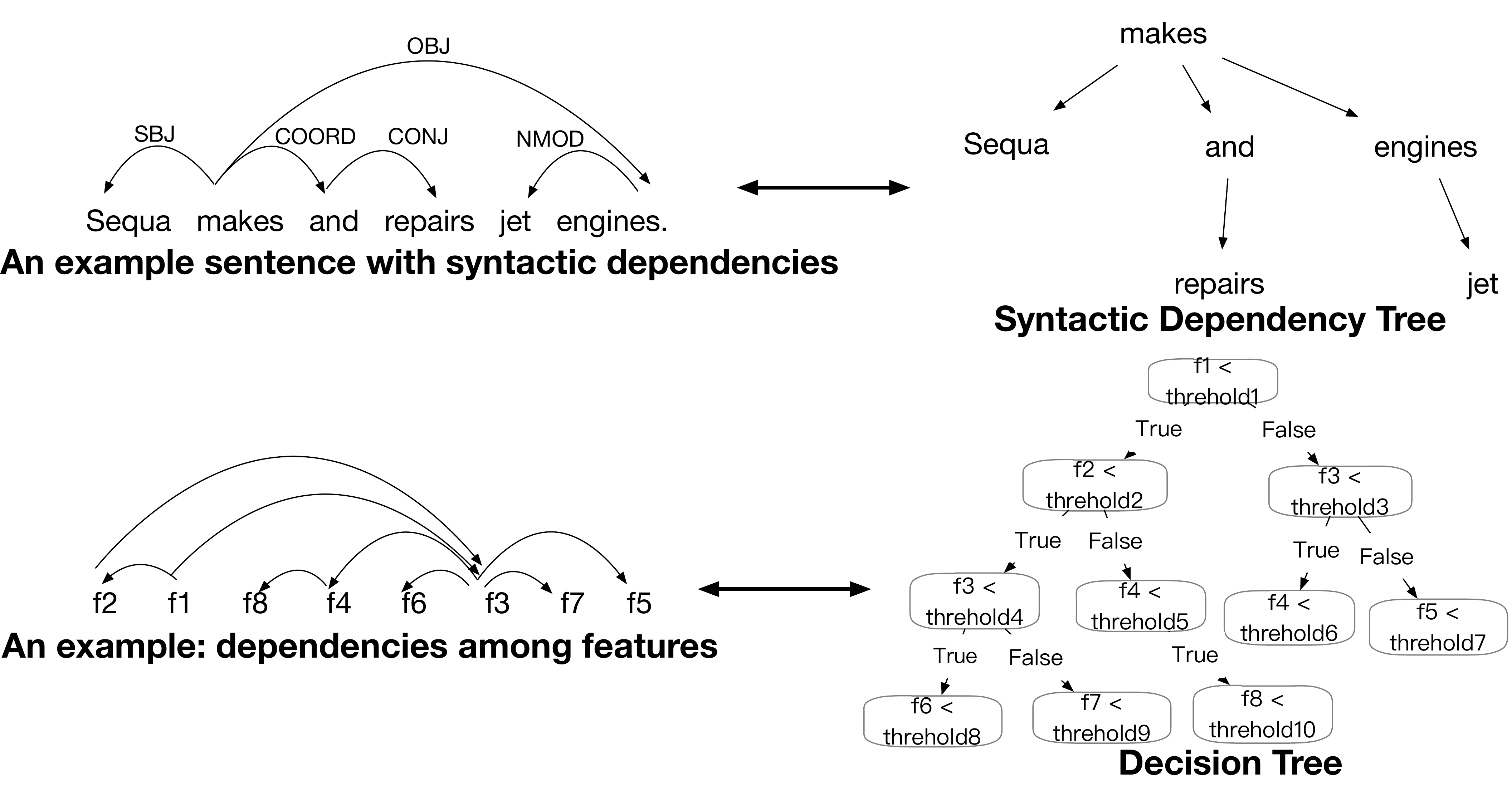}
\vspace{-1.5mm}
\caption{In NLP, syntax dependencies and syntactic dependency tree. In our framework, dependencies among features and the decision tree.}
\label{method4}
\vspace{-4mm}
\end{figure}

In the scenario of human being learning, teaching is commonly divided into several stages, such as elementary school, middle school, and university. 
For human students, in different stages, they are always taught by different teachers who have different teaching styles and different expert knowledge. 
Inspired by the human's learning process, we propose a \textbf{Hybrid Teaching} strategy, which makes agents learn from different trainers in different periods.
Figure \ref{method2} shows the general process of Hybrid Teaching strategy. Specifically, from step $0$ to step $T$, agents are  guided by one trainer.
From step $T$ to step $2T$, agents are offered advice with the help of another trainer.
Finally, after step $2T$, agents explore and learn all by themselves without the trainer. 

\begin{figure*}[h]
\centering
\includegraphics[width=16.5cm,height=7.4cm]{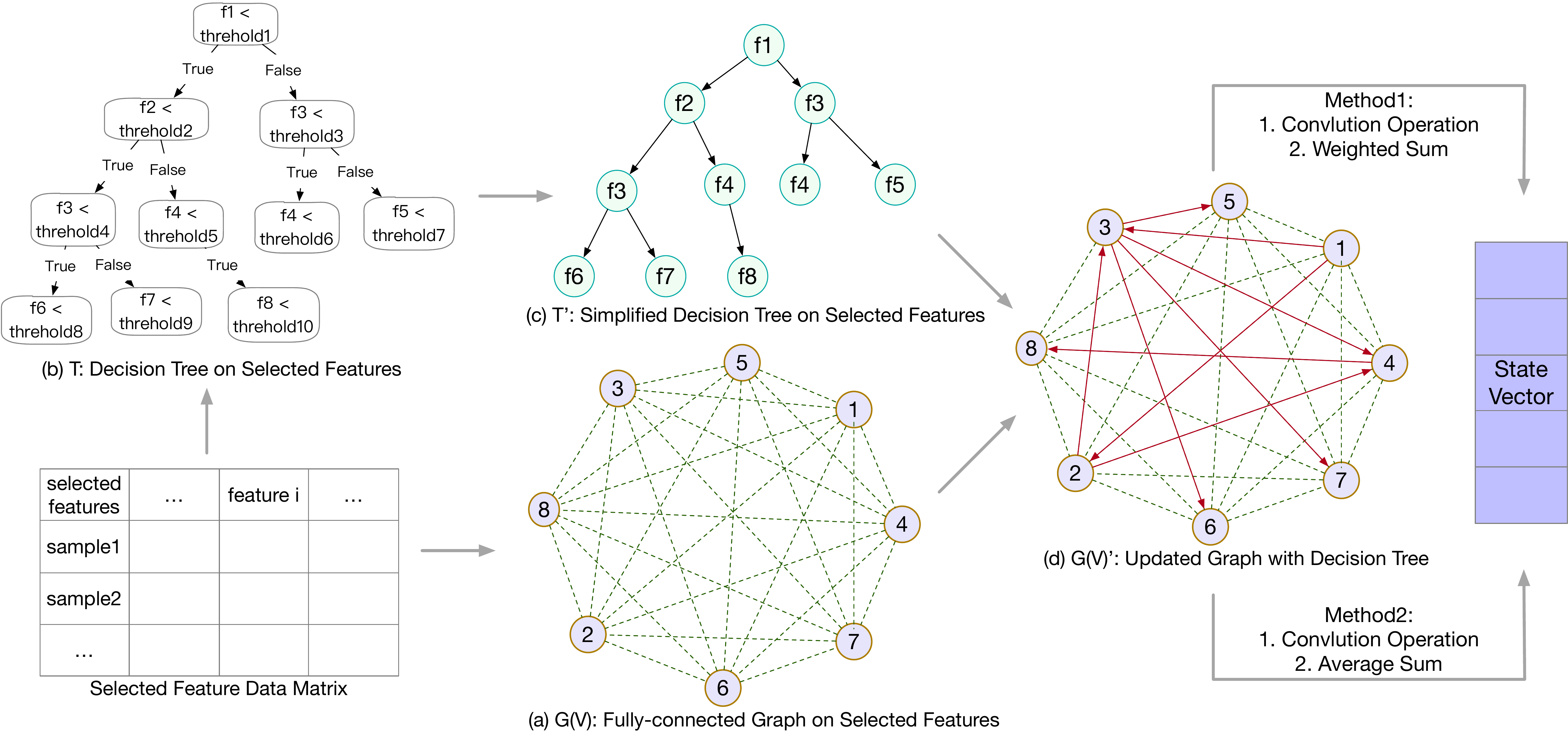}
\vspace{-4mm}
\caption{General process of state representation with decision tree structure feedback. }
\label{method3}
\vspace{-2mm}
\end{figure*}

The Hybrid Teaching strategy diversifies the teaching process and thus improves the exploration.
In the exploration, the randomness  could lead to the similar participated features. Thus, for a single trainer, it gives similar advice when taking similar participated features as input. Our strategy encourages more diversified exploration by introducing different trainers with different advice though the input is similar.
The advice given by one trainer is likely to under-perform, making agents suffer from unwise guidance. Our strategy can provide a trade-off between advice from two trainers  thus decrease the influence of bad advice.
Also, agents will lose self-study ability if they always depend on trainers' advice. Considering this, agents finally explore and learn without trainers.

\subsection{State Representation with Decision Tree Structure Feedback}
\label{sec_state_representation}

We propose a Graph Convolutional Network (GCN) based state representation method, which integrates structure feedback from the decision tree into a feature-feature correlation graph, aiming for better state representation.

Inspired by the syntactic GCN \cite{marcheggiani2017encoding,bastings2017graph} in natural language processing (NLP), we propose to apply GCN over the decision tree to represent the state in IRFS, which is similar to the GCN over the syntactic dependency tree in semantic role labeling \cite{gildea2002automatic}. 
Figure \ref{method4} shows the analogy: 
In linguistics, the syntactic dependency tree reflects syntax dependencies in a sentence, which is suited to model semantic-role structures \cite{hajivc2009conll} to produce latent feature representations of words;
similarly, the decision tree reflects the dependency relationship in a selected feature subset, which is suitable to model correlations among features better. 
Figure \ref{method3} shows the general process of state representation with tree structure feedback. We illustrate the process step by step:

\textbf{\textit{Step1}}: Following  previous work \cite{liu2019automating}, we first convert the selected features' data matrix $S$ into a fully-connected graph $G(V)$, where $V$ is the set of nodes, and every feature is represented by a node. 
For any node $n_u,n_v \in V$, we quantify the weight of edge $n_un_v$ with Pearson correlation coefficient \cite{benesty2009pearson}, denoted by $W_{u,v}$:

\begin{equation}
    W_{u,v} =  \rho_{f_u, f_v} = \frac{{\rm cov}(f_u,f_v)}{\sigma_{f_u} \sigma_{f_v}}
\end{equation}

\noindent where $f_u,f_v$ are features represented by node $n_u,n_v$; $\rho$ is Pearson correlation coefficient, $\sigma$ is standard deviation and cov is covariance.

\textbf{\textit{Step2}}: We can get a decision tree $T$ learned on the selected feature subset. We simplify the tree by only extracting the dependency relationship among features, and then get a simplified tree $T'$.

\textbf{\textit{Step3}}: For each directed edge in $T'$, we identify its corresponding edge in $G(V)$. Then, we add all the corresponding edges to $G(V)$ and get $G(V)'$. For example, in Figure \ref{method3}, for the directed edge $f_2 \to f_3$ in $T'$, its corresponding edge in $G(V)$, $n_2 \to n_3$ is added as a directed edge in $G(V)'$.

\textbf{\textit{Step4}}: To update node representation, We apply an enhanced convolution operation on $G(V)'$, which includes two different parts. One part is based on the indirect edges from feature-feature fully-connected graph; another part is based on the added directed edges from the decision tree. For each node $n_v \in V$, we denote its original representation as $h_v$, which is directly derived from $f_v$.
To better represent the state, for node $n_v$, the updated representation $h_v'$ is by:

\begin{equation}
h_v' = \lambda (\sum \limits_{n_u \in N(n_v)} W_{u,v}h_u ) + (1-\lambda)(\sum \limits_{n_w \in V}W_{w,v}h_w)
\end{equation}
where $N(n_v)$ is the set of neighbors of $n_v$, decided by the directed edges; $W_{u,v}$ is the weight of edges, in which $n_u$ is the in node and $n_v$ is the out node; { $\lambda$ is a tuning parameter. }

\textbf{\textit{Step5}}: We utilize the updated node representation to generate the representation of the selected feature subset (state). In this regard, we propose two methods to represent the state:

\noindent\textbf{Method1}. We use the weighted sum of the node representation based on the feature importance of the decision tree to represent the state. Formally, the state representation $s$ is given by: 
\begin{equation}
 s = \sum \limits_{n_v \in V} I_v h_v'
 \end{equation}
\noindent where $I_v$ is the feature importance of $f_v$ in decision tree $T$; $V$ is the set of nodes. 

\noindent\textbf{Method2}. We use the average sum of the node representation to represent the state. Formally, the state representation $s$ is given by: 
\begin{equation}
 s = \sum \limits_{n_v \in V} \frac{1}{|V|} h_v' 
\end{equation}
\noindent where $h_v'$ is the updated representation of $n_v$, and $V$ is the set of nodes.

\begin{figure}[h]
\centering
\includegraphics[width=7cm]{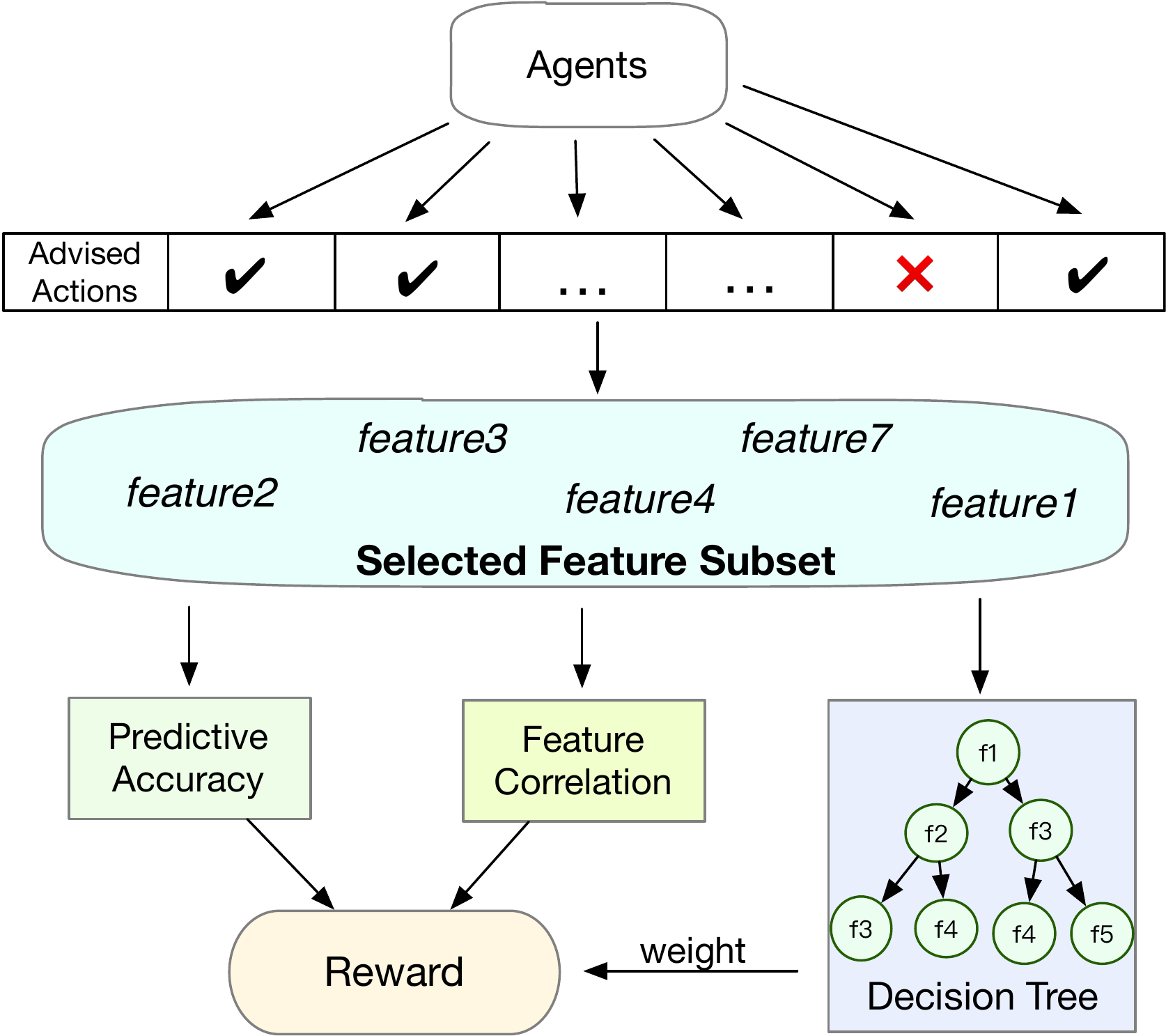}
\caption{General process of measuring reward with feature importance in the decision tree.}
\vspace{-2mm}
\label{method6}
\end{figure}

\subsection{Personalized Reward Schemes}

\label{method_reward_part}

In reinforcement learning, precise reward measurement is important to exploration and evaluation. To improve the way to calculate reward, we propose two personalized reward schemes: (i) to personalize reward based on the structured feedback from decision tree, and (ii) to personalize reward based on the historical action records.
These two schemes consider the predictive accuracy $Acc$ of downstream task and the feature correlation $R$ of the selected feature subset for reward measurement. 


\noindent \textbf{Predictive Accuracy}. We aim to find an optimal feature subset which gets good performance in the downstream task. Naturally, we propose to utilize the performance $Acc$ as part of reward. The higher the performance is, the higher reward agents should receive.

\noindent \textbf{Feature Correlation}. We also propose to use another characteristic of the selected feature subset to measure the reward: feature correlation. Specifically, a qualified feature subset is usually of low average correlation; high average correlation of features is unfavorable for an optimal feature subset.

\subsubsection{Measuring Reward with Decision Tree Structured Feedback}

\label{section_reward-scheme1}

In automated feature selection, reinforcement learning agents select features into the subset in each step. Intuitively, different features in the selected subset have different importance, and the action to select more important features should receive higher reward. The structure of a decision tree could provide a measurement of the importance of features. Considering this, we propose to personalize the reward with the feature importance from the decision tree  learned on the selected feature subset. Figure \ref{method6} shows the general process. We detail the steps as follows:

\textit{\textbf{Step1}}: Given agent actions $\{a_1^t, a_2^t, ..., a_N^t\}$ at step $t$, we get a selected feature subset $F_{s} = \{ f_i\;|\;a_i^t = 1 \}$. Then, we calculate predictive accuracy $Acc$ of the downstream task on the selected features.

\textit{\textbf{Step}2}: We use Pearson correlation coefficient \cite{benesty2009pearson} to quantify the correlation of selected feature subset. 
The feature correlation, denoted by $R$, is computed as the sum of pairwise Pearson correlation coefficient. Formally:
\begin{equation}
    \centering
 R = \frac{1}{ |F_{s}|^2} \sum_{ f_u, f_v \in F_{s}} \rho_{f_u,f_v} 
\end{equation}

\textit{\textbf{Step2}}: At step $t$, we train a decision tree $T^t$ on the current selected feature subset. Then, we can get the importance of each feature from the decision tree. For feature $f_i$ at step $t$, its feature importance is denoted by $I_i^t$.

\begin{figure}[h]
\centering
\includegraphics[width=9cm]{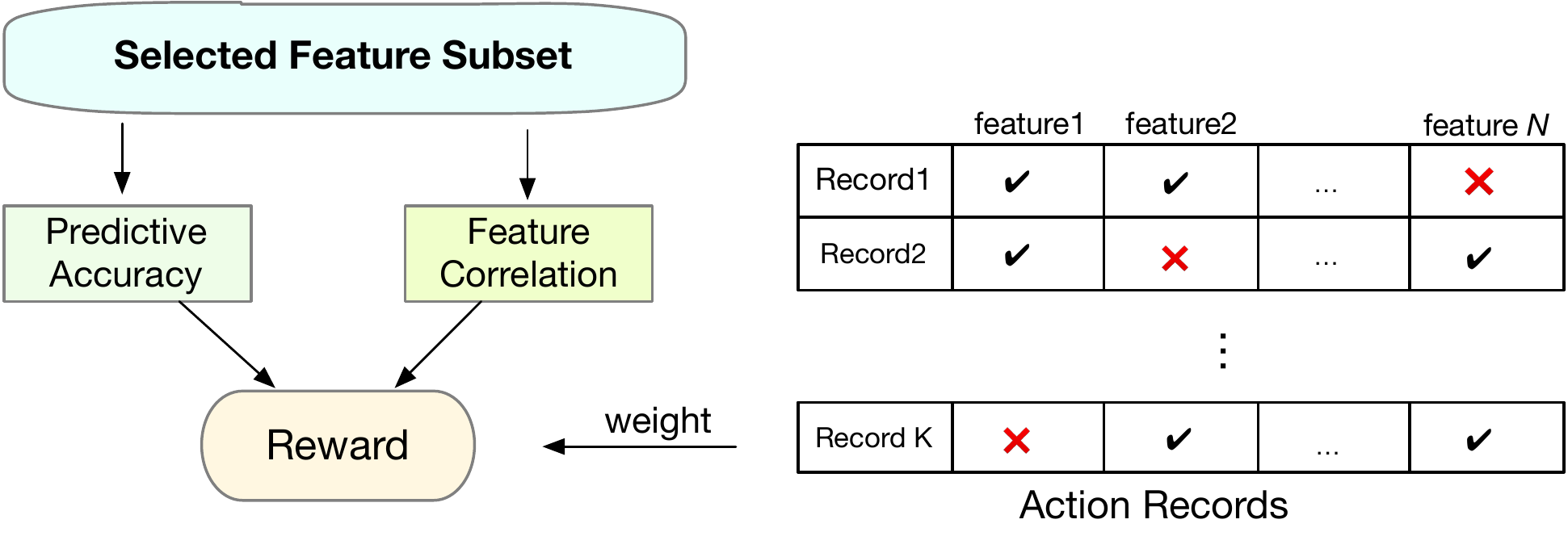}
\caption{General process of measuring reward with historical action records.}
\vspace{-2mm}
\label{method7}
\end{figure}

\textit{\textbf{Step3}}: We use the feature importance to weight the reward assigned to agents which are responsible for the selection of different features. Formally, for agent $agt_i$ at step $t$, its reward is computed by:

\begin{equation}
\centering
 r_i^t=\left\{
\begin{aligned}
 I_i^t(Acc - \beta\,R), a_i^t = 1  \\
0, a_i^t = 0 \\
\end{aligned}
\right.
\end{equation}
where $\beta$ is a tuning parameter, $Acc$ is predictive accuracy, and $R$ is feature correlation.

\subsubsection{Measuring Reward with Historical Action records}
\label{section_reward_scheme2}

In reinforcement learning, agents are more likely to choose advantageous actions for the purpose of maximizing the long-term reward. Intuitively, if a feature is always selected by the agent, this feature could be considered important to the optimal feature subset. Thus, the selection of important features should receive higher reward. In this regard, we propose to personalize the reward towards agents based on the importance of their corresponding features, which can be quantified by  their selected  frequency  ratio. Figure \ref{method7} shows the general process of measuring reward. We introduce the steps of this reward scheme as follows:

\textit{\textbf{Step1}}: Similar to Section \ref{section_reward-scheme1}, we firstly calculate predictive accuracy $Acc$ of the downstream task, and then quantify the feature correlation $R$ of the selected feature subset.

\textit{\textbf{Step2}}: We record actions and denote historical action records by $ \{m_1^t, m_2^t, ...,m_N^t\}$, where $m_i^t = (a_i^0, a_i^1, ..., a_i^t)$ is for $agt_i$ at step $t$.
Next, we use action record to calculate the importance of every feature, which is used to weight reward towards agents. For agent $agt_i$ at step $t$, its reward weight is denoted by $W_i^t$:
\begin{equation}
    \centering
 W_i^t = \frac{ \sum m_i^t}{ \sum_{i=1}^{N} \sum m_i^t} 
\end{equation}

\textit{\textbf{Step3}}:, we measure the reward by combining the predictive accuracy, the feature correlation and the action records. Formally, for agent $agt_i$ at step $t$, its reward is measured by:
\begin{equation}
\centering
 r_i^t=\left\{
\begin{aligned}
 W_i^t(Acc - \beta\,R), a_i^t = 1  \\
0, a_i^t = 0 \\
\end{aligned}
\right.
\end{equation}
where $\beta$ is a tuning parameter, $Acc$ is predictive accuracy, and $R$ is feature correlation.

\vspace{-0.1cm}
\section{Experiment}
\vspace{-0.1cm}
We evaluate the proposed methods in feature selection with  different real-world datasets. Table \ref{dataset} shows the description of the datasets.

\vspace{-3mm}
\subsection{Data Description}

\noindent\textbf{Pen-based Recognition of Digits (PRD) Dataset}. This dataset creates a digit database by collecting samples from different writers. In the experiments, 10992 samples written by 30 writers are split for training and  cross-validation. All input attributes are integers in the range from 0 to 100. The label is from 0 to 9.  \cite{Dua:2019}.

\noindent\textbf{Forest Cover (FC) Dataset}. We select a publicly available dataset from Kaggle \footnote{\url{ https://www.kaggle.com/c/forest-cover-type-prediction/data} }. This dataset includes 15120 samples and 54 different characteristics of wilderness areas, which are used to predict forest cover type. The class labels (forest cover type values) are from 1 to 7.

\noindent\textbf{Spam Dataset}. This dataset is a collection of spam emails which came from the postmaster and individuals who had filed spam. This dataset includes 4601 samples and 57 different features, most of which indicate whether a particular word or character was frequently occurring in the email. The class label is 1 (spam) or 0 (not spam). \cite{Dua:2019}.

\noindent\textbf{Insurance Company Benchmark (ICB) Dataset}. This dataset contains information about customers, which consists of 86 variables and includes product usage data and socio-demographic data derived from zip area codes \cite{van2000coil}.

\noindent\textbf{Nomao Dataset}. This dataset has 34465 instances and 120 attributes, which consists of 89 continuous attributes and 31 nominal ones (including the attributes `label' and `id') \cite{candillier2012design}.

\begin{table}
\footnotesize
\begin{center} 
\caption{Description of the dataset.}
\label{dataset}
\begin{tabular}{|p{1.2cm}|p{1.2cm}|p{1.2cm}|p{1.2cm}|p{1.2cm}|} 
\hline  
\textbf{Dataset} &\textbf{PRD}    &\textbf{FC}   &\textbf{Spam}  &\textbf{ICB}    \\\hline  
Features& 16&   54&    57  &   86 \\  \hline  
Samples& 10992&   15120&    4601  &   5000 \\  \hline  
\textbf{Dataset} & \textbf{Nomao} &\textbf{Musk} &\textbf{ESR} &\textbf{QSAR  } \\ \hline
Features& 120&   168&    178  &   1024 \\ \hline
Samples&  34465&   6598&    11500  &   8992 \\  \hline

\end{tabular} 
\end{center}
\vspace{-4mm}
\end{table}

\noindent\textbf{Musk Dataset}. This dataset describes a set of 102 molecules of which 39 are judged by human experts to be musks and the remaining 63 are judged to be non-musks. It includes 6598 samples and the aim is to classify the label `musk' and `non-musk' \cite{Dua:2019}.

\begin{figure*}[htbp]
\centering
\subfigure[PRD Dataset]{
\includegraphics[width=4.1cm]{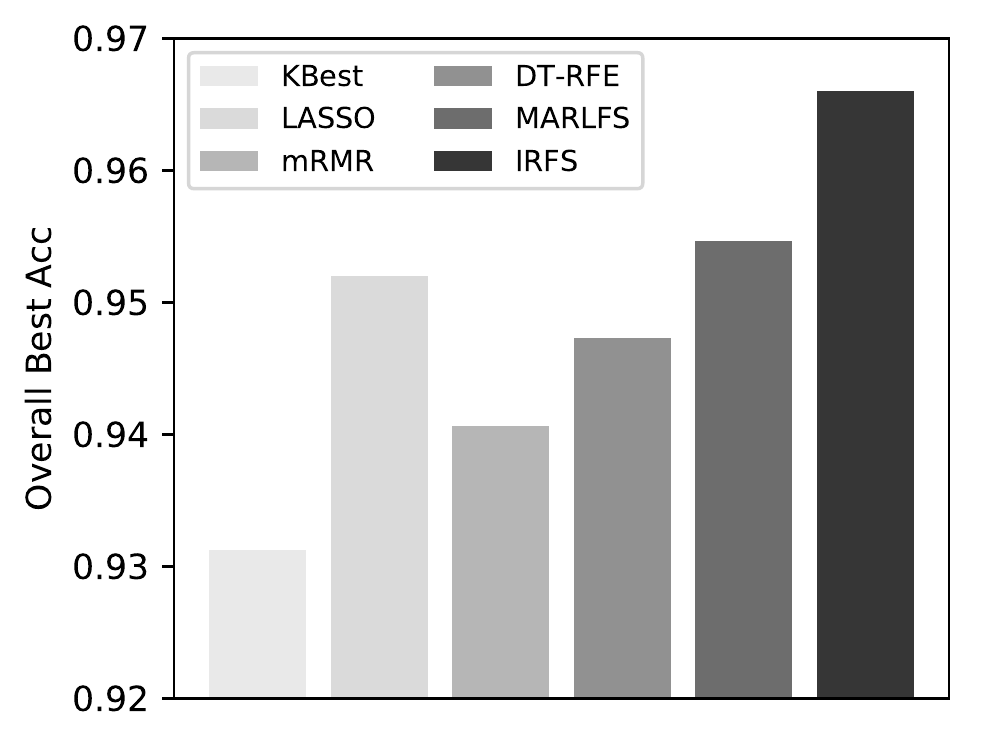}
}
\hspace{-0mm}
\subfigure[FC Dataset]{
\includegraphics[width=4.1cm]{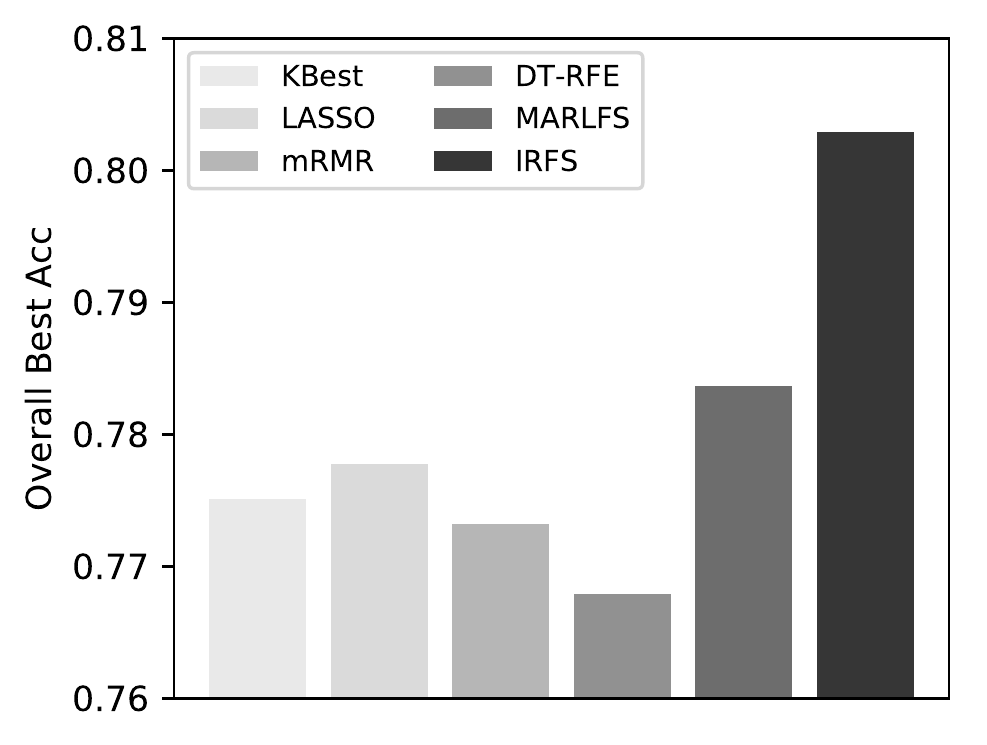}
}
\hspace{-0mm}
\subfigure[Spam Dataset]{
\includegraphics[width=4.1cm]{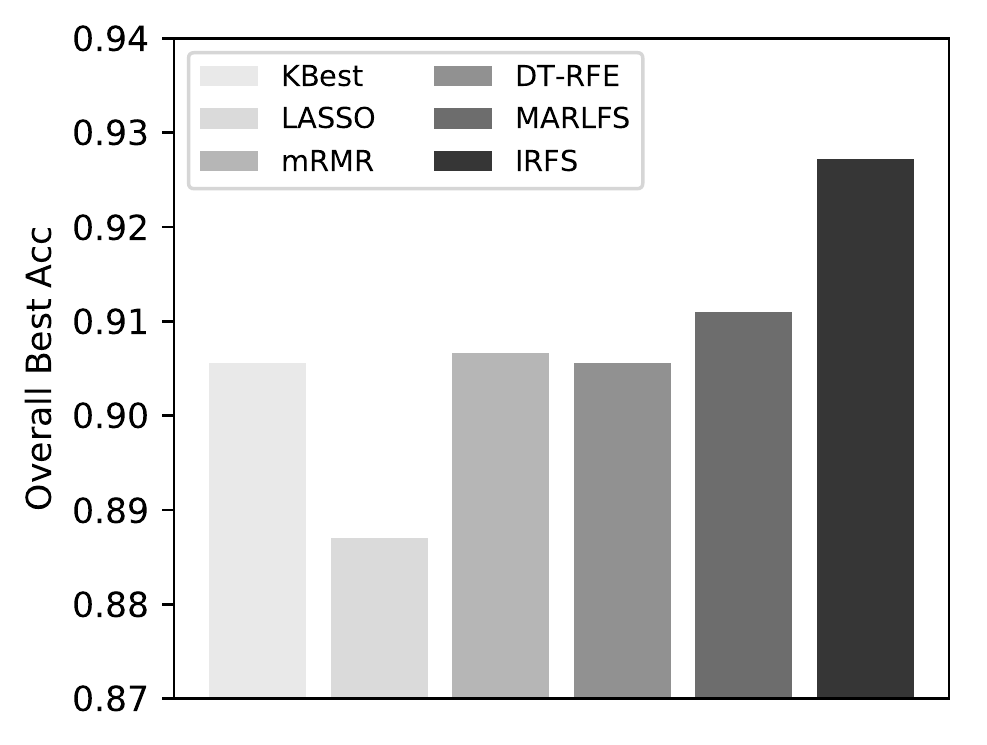}
}
\hspace{-0mm}
\subfigure[ICB Dataset]{
\includegraphics[width=4.1cm]{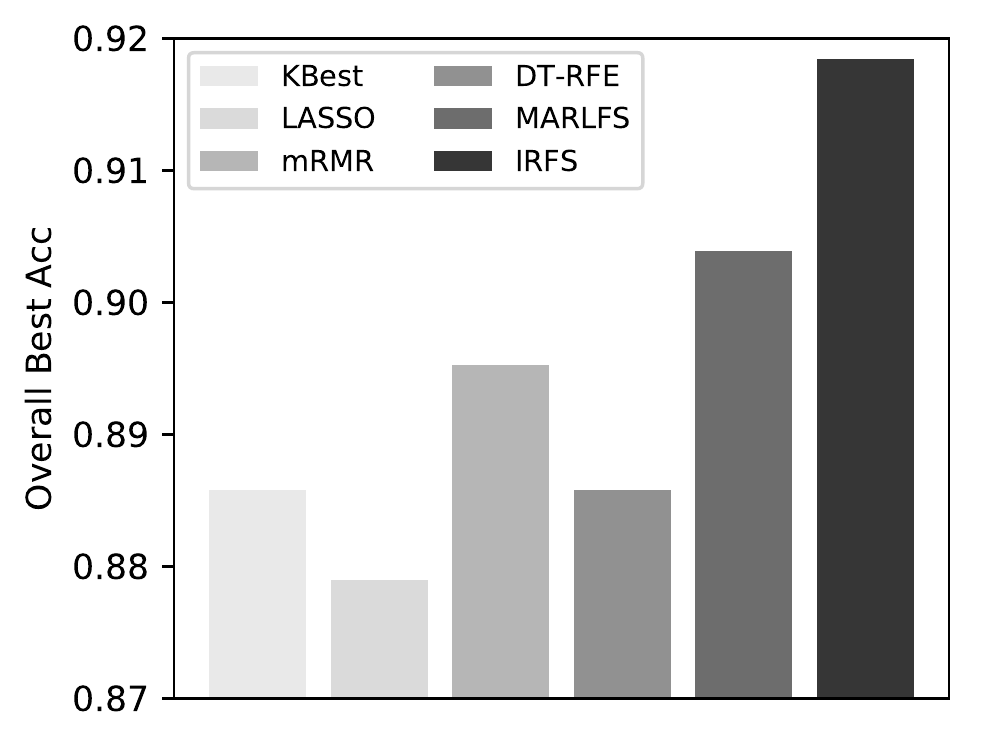}
}
\subfigure[Nomao Dataset]{
\includegraphics[width=4.1cm]{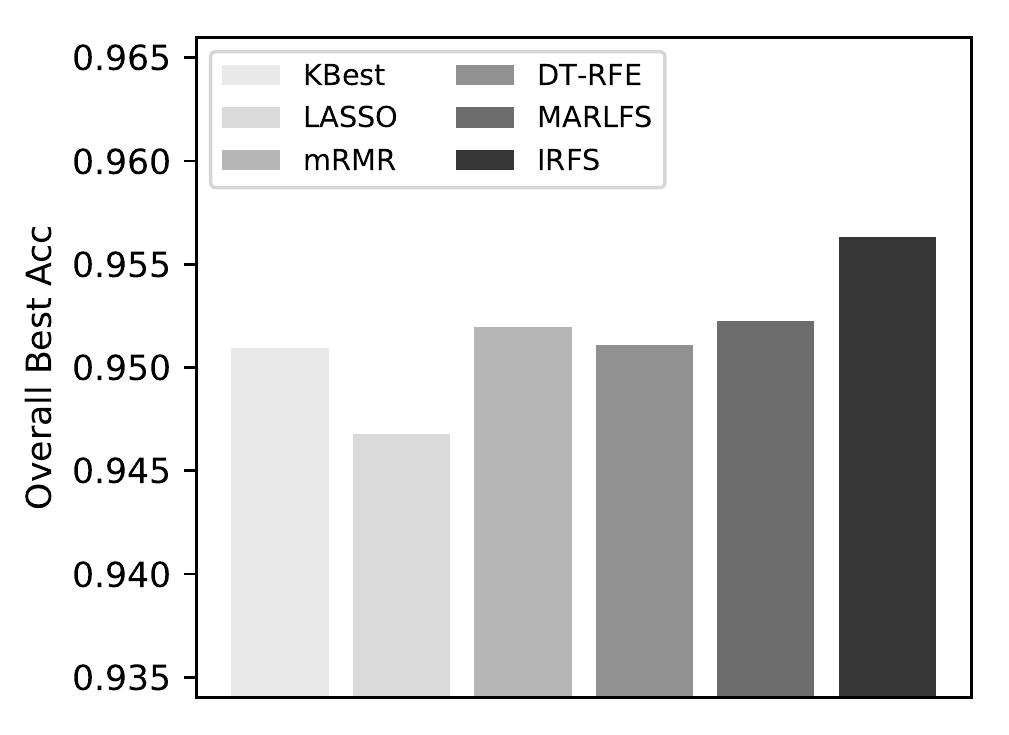}
}
\hspace{-0mm}
\subfigure[Musk Dataset]{
\includegraphics[width=4.1cm]{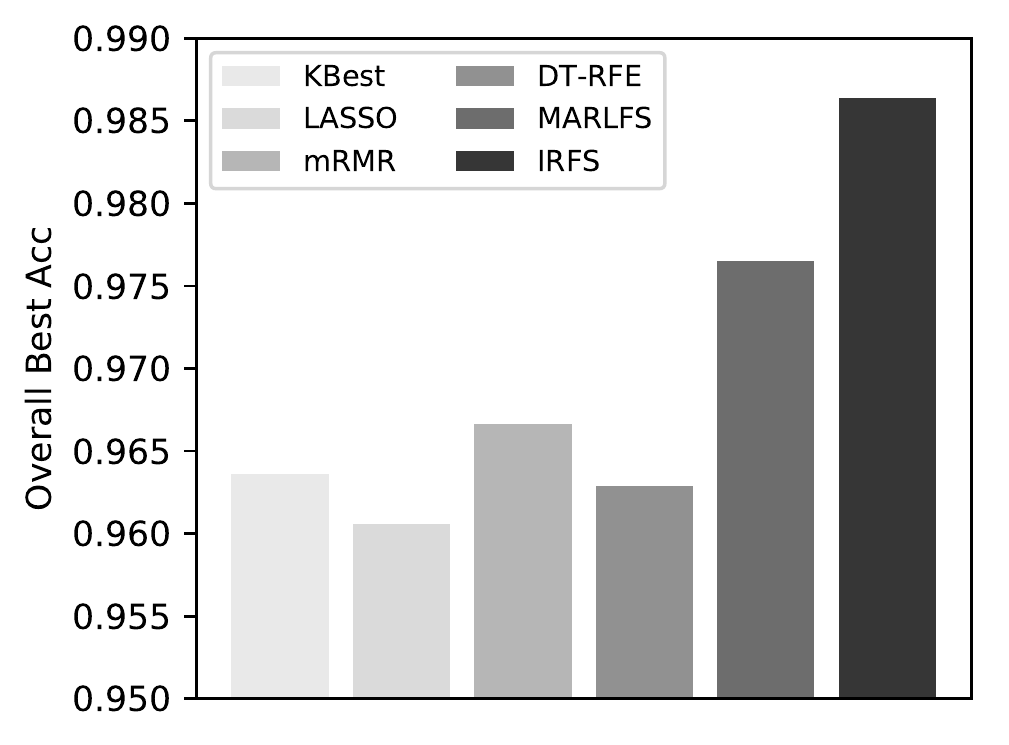}
}
\hspace{-0mm}
\subfigure[ESR Dataset]{
\includegraphics[width=4.1cm]{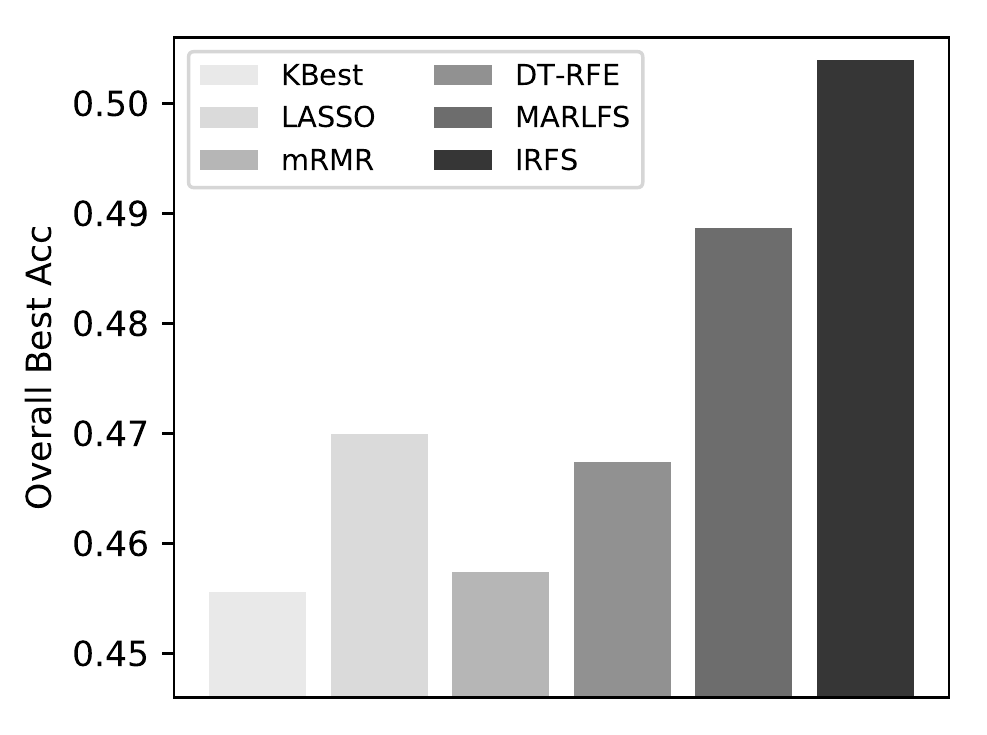}
}
\hspace{-0mm}
\subfigure[QSAR Dataset]{
\includegraphics[width=4.1cm]{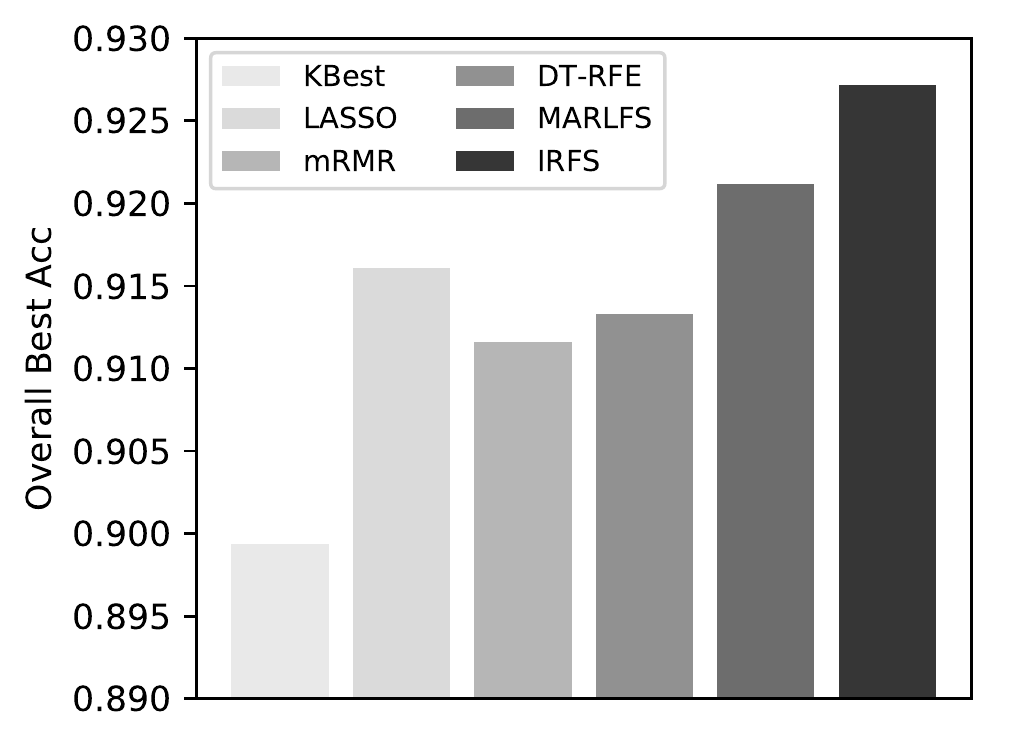}
}

\vspace{-2mm}
\caption{Overall Best Acc of different feature selection algorithms. }
\label{exp_pic1}
\vspace{-4mm}
\end{figure*}

\noindent\textbf{Epileptic Seizure Recognition (ESR) Dataset}. This dataset contains 178 attributes and 11500 samples, with the class label ranging from 1 to 5. All subjects falling in classes 2, 3, 4, and 5 are subjects who did not have epileptic seizure. Only subjects in class 1 have epileptic seizure. \cite{andrzejak2001indications}

\noindent\textbf{QSAR oral toxicity Dataset }. This dataset includes 8992 instances and 1024 attribute, which is used to develop classification QSAR models for the discrimination of very toxic/positive (741) and not very toxic/negative (8251) molecules \cite{ballabio2019integrated}.

\vspace{-3mm}
\subsection{Evaluation Metrics}

We use the following metrics for evaluation, in order to show the performance of our proposed methods.

\textbf{Best Acc}. Accuracy is the ratio of the number of correct predictions to the number of all predictions. Formally, the accuracy is given by $Acc = \frac{TP+TN}{TP+TN+FP+FN}$, where $TP,TN, FP, FN$ are true positive, true negative, false positive and false negative for all classes. In feature selection, an important task is to find the optimal feature subset, which has good performance in the downstream task. Considering this, we use Best Acc (BA) to stand for the performance of feature selection, which is given by $BA_{l} = max(Acc_i, Acc_{i+1}, ..., Acc_{
i+l})$, where $i$ is the beginning step and $l$ is the number of exploration steps. 

\textbf{Ave Acc}. Due to the randomness of the exploration, the performance of the selected feature subset in downstream task may vary significantly from step to step, which makes it difficult to measure the performance of a certain period of time. As a result, we calculate the Average Accuracy (Ave Acc) to show the average performance of the reinforced feature selection. Formally, the Ave Acc (AA) is given by: $AA_{l} = \frac{Acc_i+Acc_{i+1}+ ... + Acc_{i+l}}{l}$, where $i$ is the beginning step and $l$ is the number of exploration steps.

\vspace{-3mm}
\subsection{Baseline Algorithm}

We compare the feature selection performance of our proposed method with the following five baseline algorithms, where algorithm (1)--(4) are traditional feature selection methods, and algorithm (5) is the reinforced feature selection method. 

(1) K-Best Feature Selection. This algorithm \cite{yang1997comparative} ranks features by their ranking scores with the label vector and selects the top $k$ highest scoring features. In the experiments, we set $k$ equals to half of the number of input features.

(2) Decision Tree Recursive Feature Elimination (DT-RFE). RFE \cite{granitto2006recursive} selects features by  selecting smaller and smaller feature subsets until finding the subset with certain number of features. DT-RFE first trains a decision tree on all the features to get feature importance; then it recursively deselects the least important features. In the experiments, we set the selected feature number half of the feature space.

(3) mRMR. The mRMR \cite{peng2005feature} ranks features by minimizing feature’s redundancy and maximizing their relevance in the meantime. Then, it selects the top-$k$ ranked features as the feature selection result. In experiments, we set $k$ equals to half of the number of input features.

(4) LASSO. LASSO \cite{tibshirani1996regression} conducts feature selection and shrinkage via $l1$ penalty. In this method, features whose coefficients are 0 will be dropped. All the parameter settings are the same as \cite{liu2019automating}.

(5) Multi-agent Reinforcement Learning Feature Selection (MARLFS). MARLFS \cite{liu2019automating} is a basic multi-agent reinforced feature selection method, which could be seen as a variant of our IRFS without any trainer. To compare fairly, the downstream task is set the same as our framework.

In the experiments, our KBest based trainer uses mutual information to select features, supported by scikit-learn. 
Our Decision Tree based teacher uses decision tree classifier with default parameters in scikit-learn \footnote{\url{https://scikit-learn.org/stable/modules/tree.html}}. 
In state representation, following previous work \cite{liu2019automating}, we utilize graph convolutional network (GCN) \cite{kipf2016semi} to update features, where features are fully connected in a complete feature graph. In the experience replay \cite{lin1992self,mnih2015human}, each agent has its memory unit. For agent $agt_i$ at step $t$, we store a tuple $\{s_i^t, r_i^t, s_i^{t+1}, a_i^t\}$ to the memory unit. 
The deep-Q network of agents is set to  two linear layers of 128 middle
states with ReLU as activation function. In the exploration process, the discount factor $\gamma$ is set to 0.9, and we use $\epsilon$-greedy exploration with $\epsilon$ equals to 0.9. To train the policy networks, we select mini-batches with 16 as batch size and
use Adam Optimizer with a learning rate of 0.01.
To compare fairly with baseline algorithms, we fix the downstream task as a decision tree classifier with default parameters in scikit-learn. We randomly split the data into train data ($80\%$) and test data ($20\%$). All the evaluations are performed on Intel E5-1680 3.40GHz CPU in a x64 machine, whose RAM is 128GB and operation system is CentOS 7.4.  

\vspace{-3mm}
\subsection{Overall Performance}

\begin{figure*}[htbp]
\centering
\subfigure[PRD Dataset]{
\includegraphics[width=4.4cm]{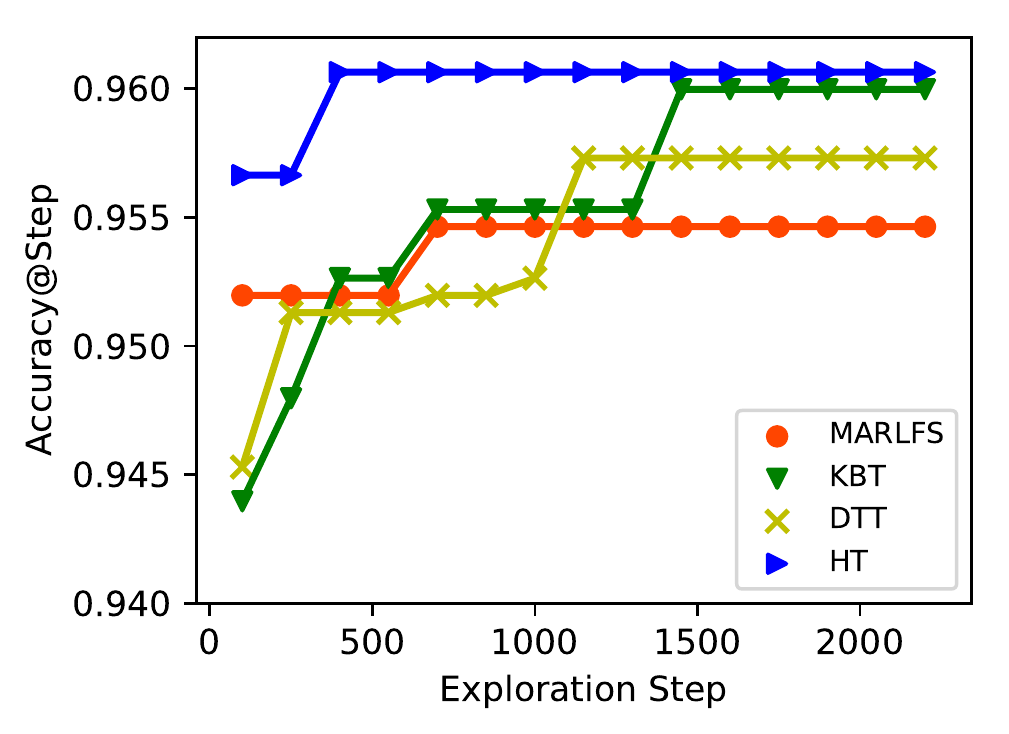}
}
\hspace{-4mm}
\subfigure[FC Dataset]{
\includegraphics[width=4.3cm]{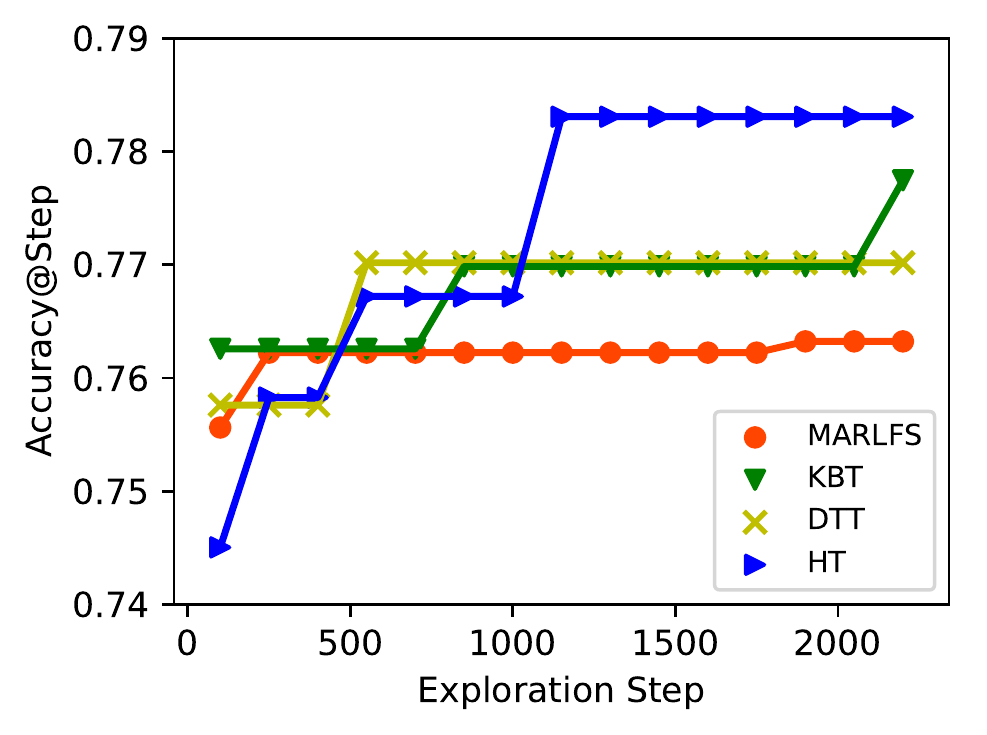}
}
\hspace{-4mm}
\subfigure[Spam Dataset]{
\includegraphics[width=4.4cm]{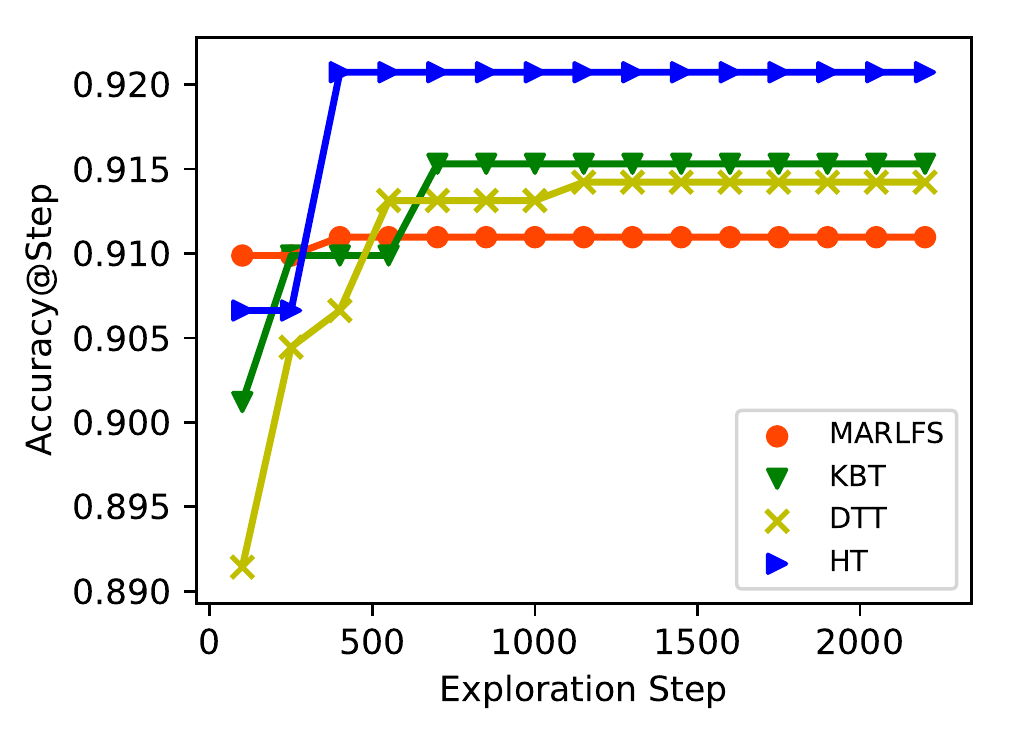}
}
\hspace{-4mm}
\subfigure[ICB Dataset]{
\includegraphics[width=4.4cm]{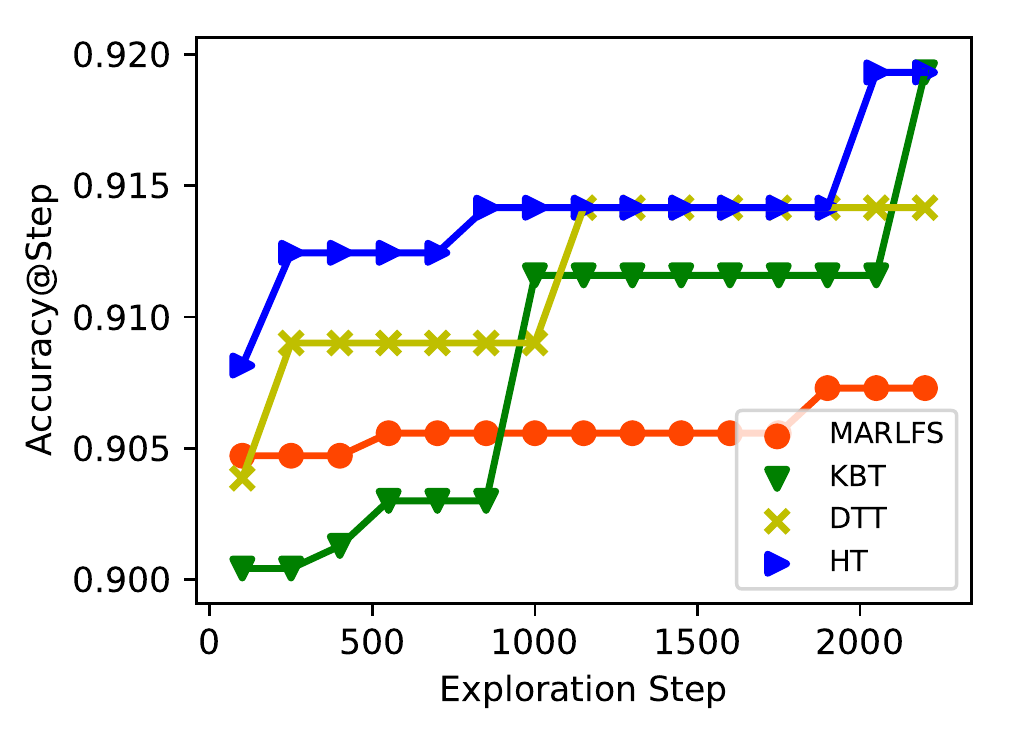}
}
\hspace{-4mm}
\subfigure[Nomao Dataset]{
\includegraphics[width=4.4cm]{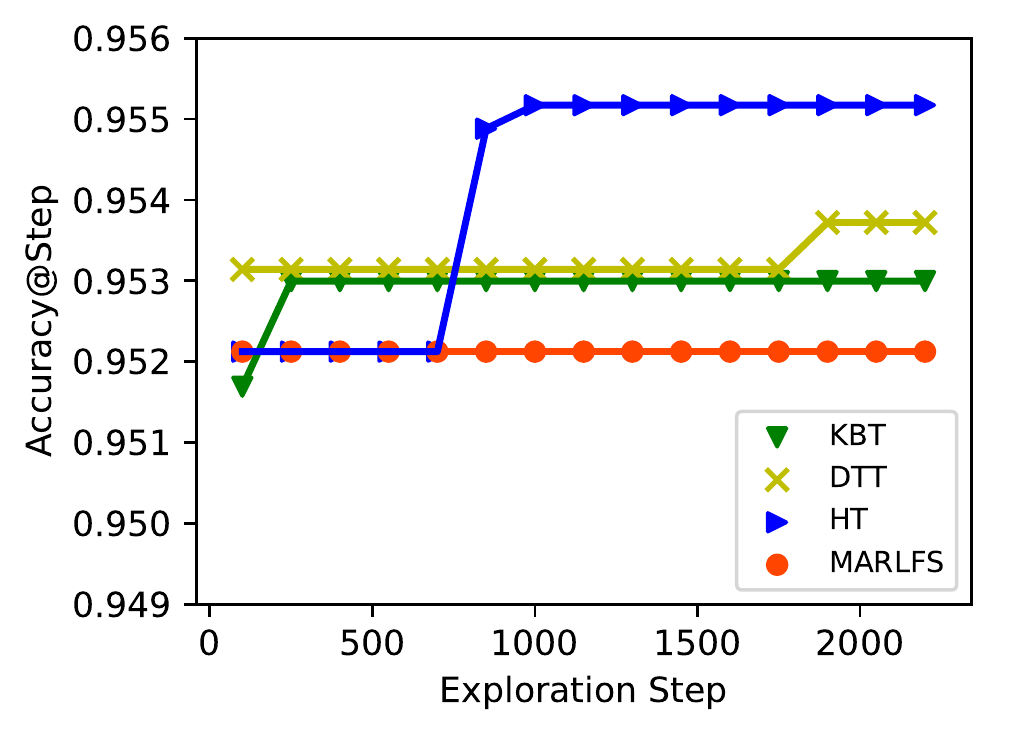}
}
\hspace{-4mm}
\subfigure[Musk Dataset]{
\includegraphics[width=4.4cm]{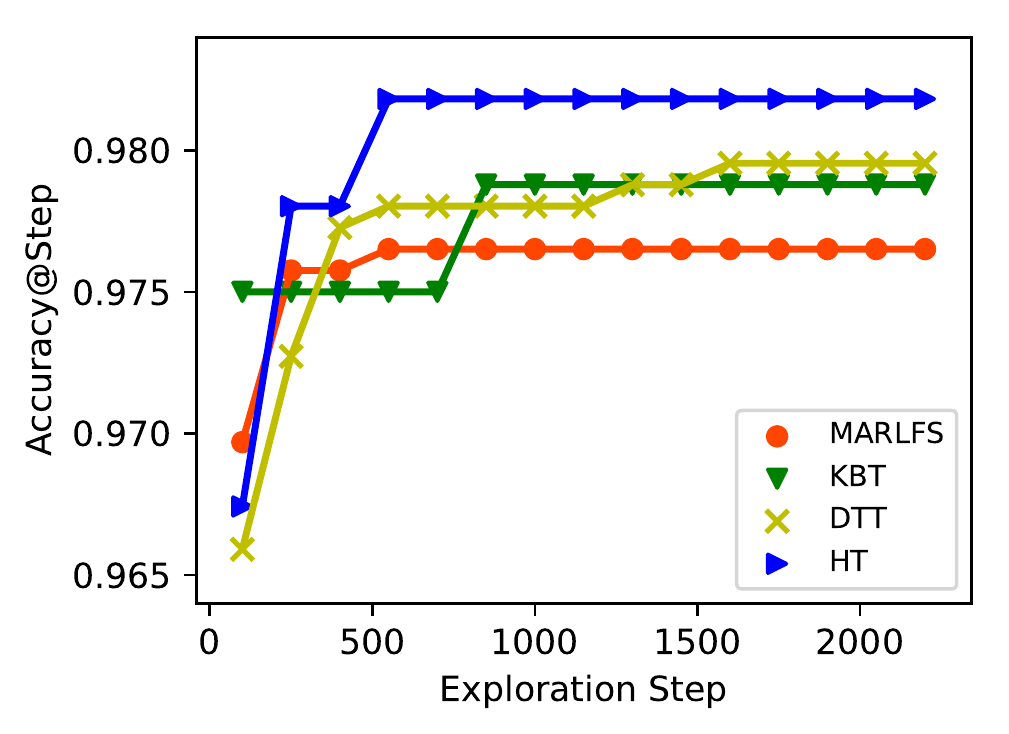}
}
\hspace{-4mm}
\subfigure[ESR Dataset]{
\includegraphics[width=4.4cm]{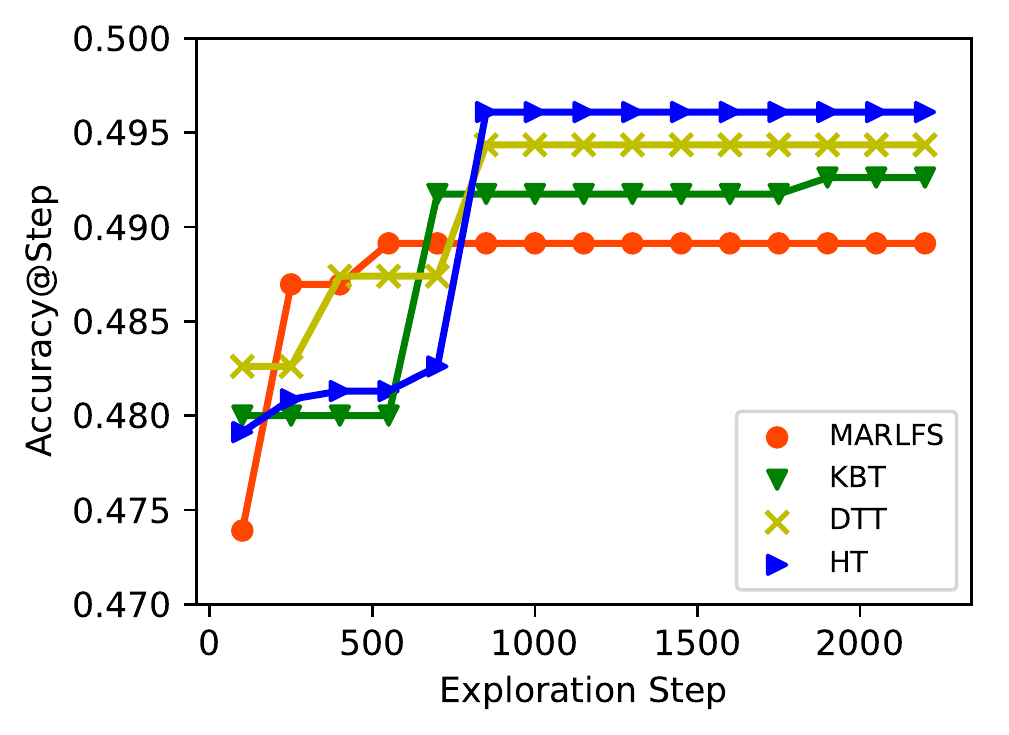}
}
\hspace{-4mm}
\subfigure[QSAR Dataset]{
\includegraphics[width=4.5cm]{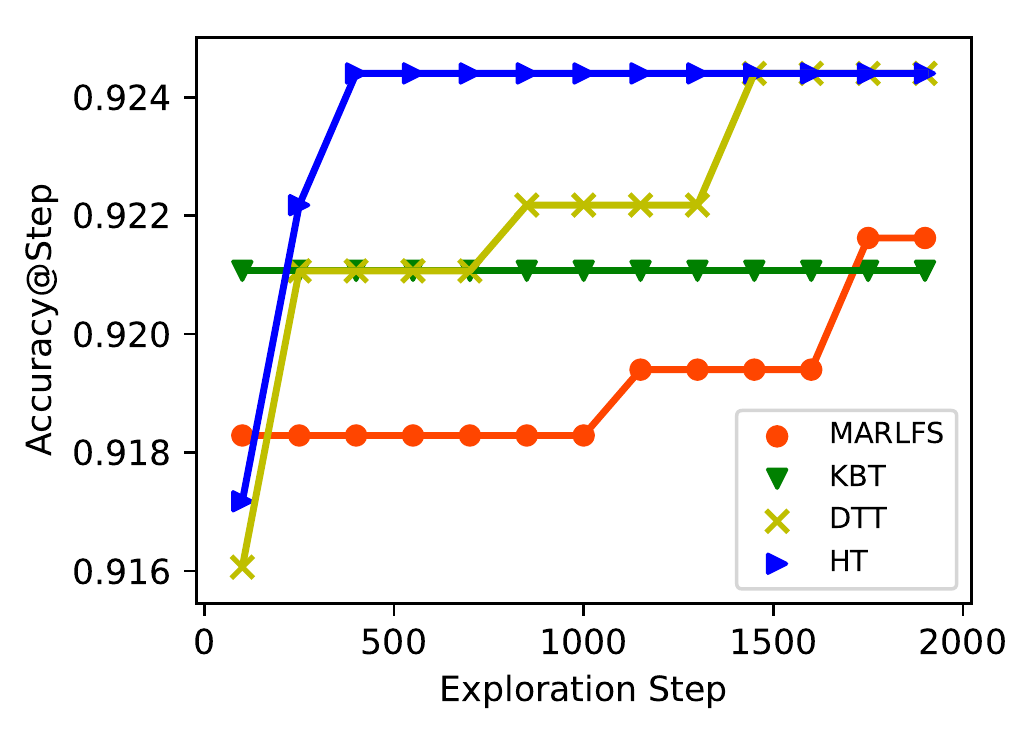}
}
\vspace{-2mm}
\caption{Exploration efficiency comparison of IRFS methods.}
\label{exp_pic2}
\vspace{-5mm}
\end{figure*}

\begin{figure*}[htbp]
\centering
\subfigure[PRD Dataset]{
\includegraphics[width=4.15cm]{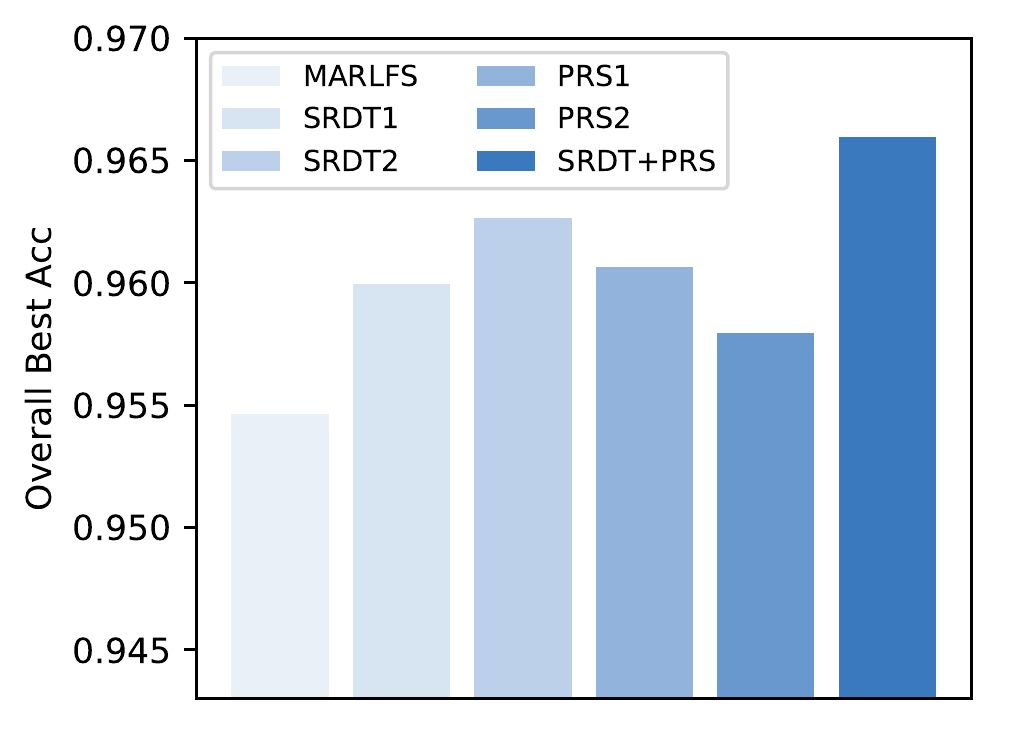}
}
\hspace{-0mm}
\subfigure[FC Dataset]{
\includegraphics[width=4.0cm]{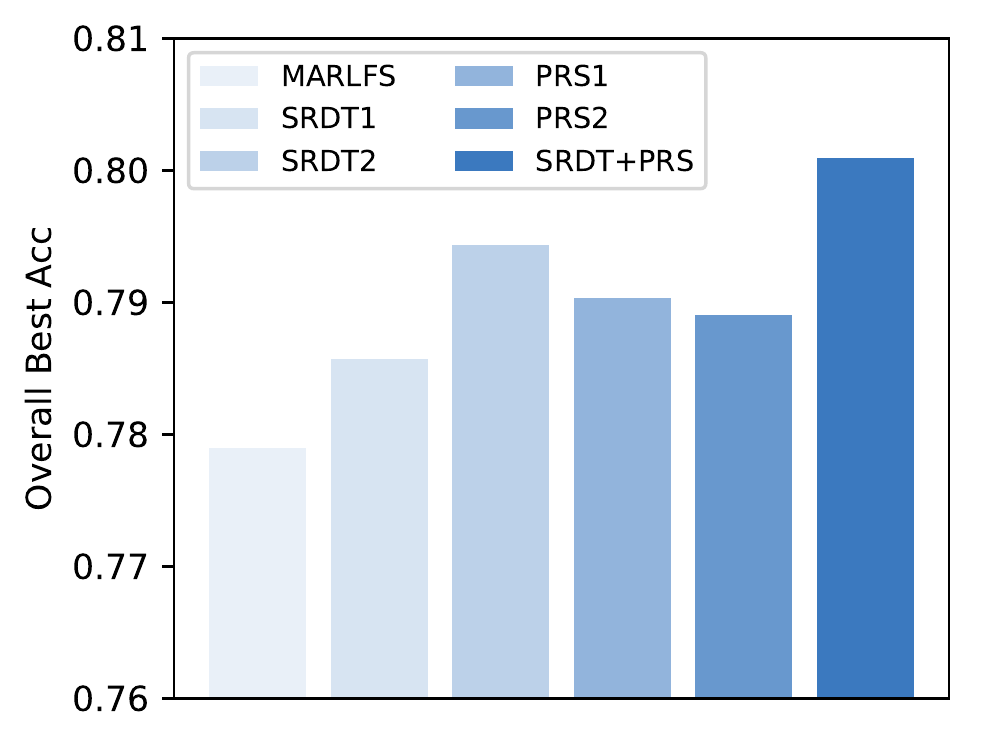}
}
\hspace{-0mm}
\subfigure[Spam Dataset]{
\includegraphics[width=4.1cm]{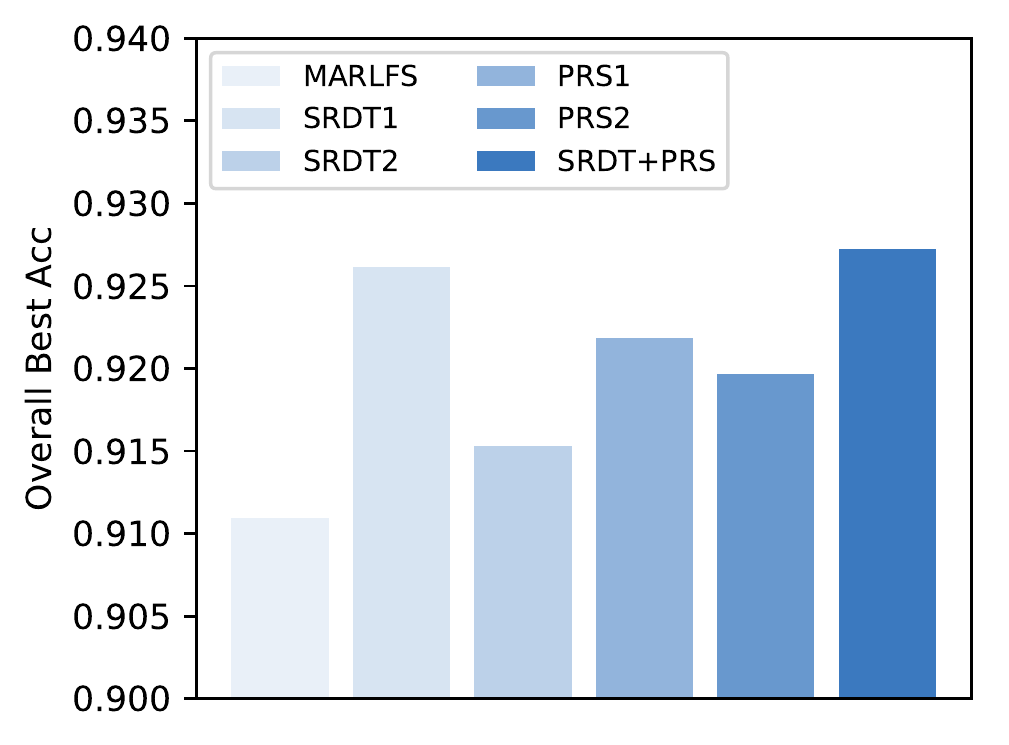}
}
\hspace{-0mm}
\subfigure[ICB Dataset]{
\includegraphics[width=4.1cm]{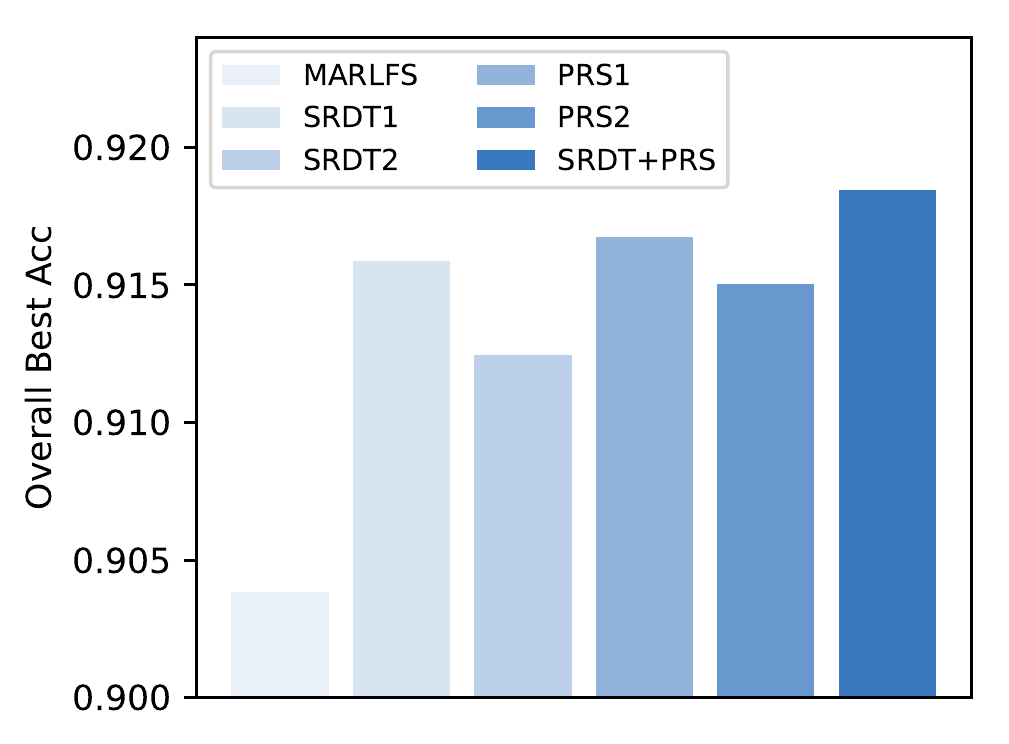}
}
\hspace{-0mm}
\subfigure[Nomao Dataset]{
\includegraphics[width=4.1cm]{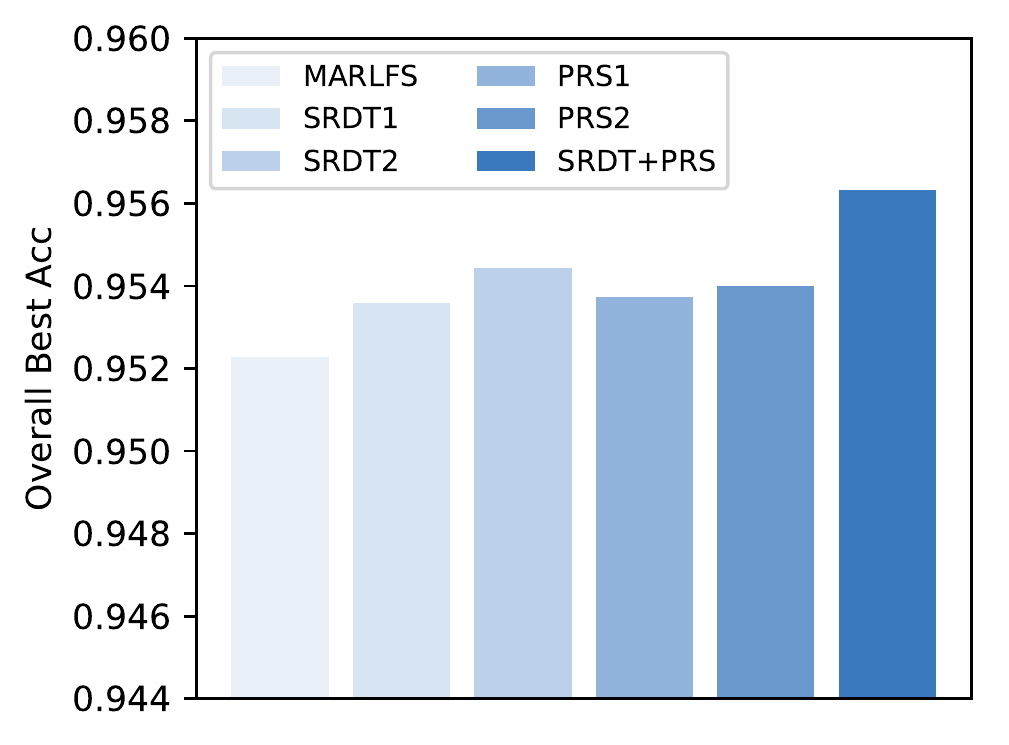}
}
\hspace{-0mm}
\subfigure[Musk Dataset]{
\includegraphics[width=4.15cm]{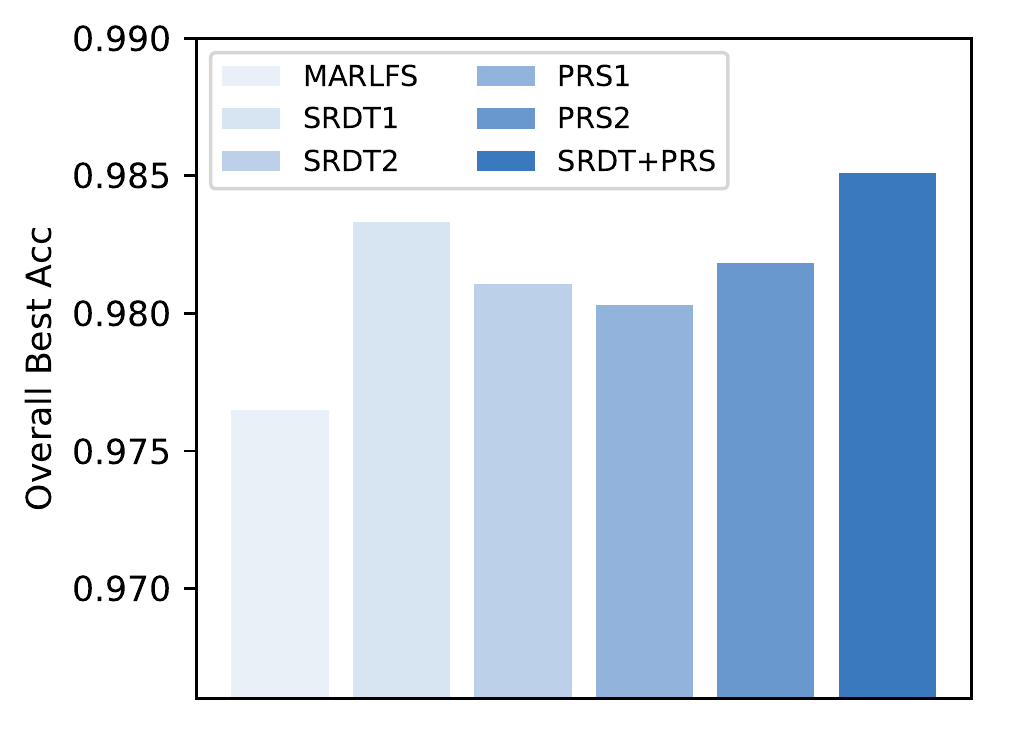}
}
\hspace{-0mm}
\subfigure[ESR Dataset]{
\includegraphics[width=4.05cm]{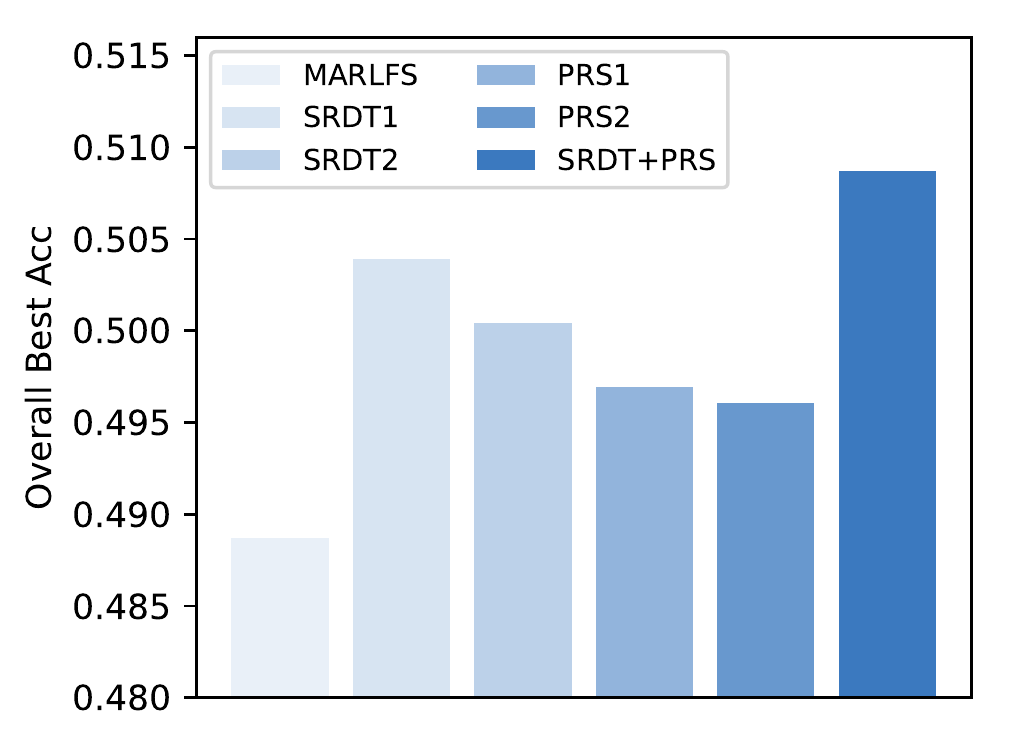}
}
\hspace{-0mm}
\subfigure[QSAR Dataset]{
\includegraphics[width=4.15cm]{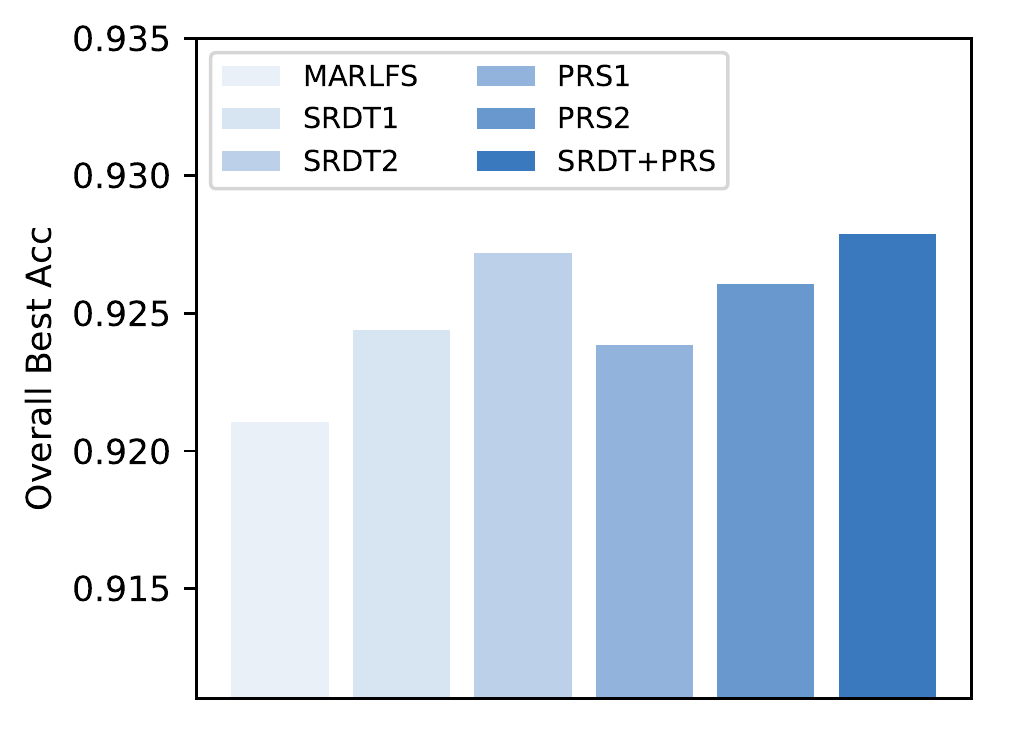}
}
\vspace{-2mm}
\caption{Best Acc comparison of variant methods. }
\label{exp_pic3}
\vspace{-5mm}
\end{figure*}

We compare our proposed method with baseline methods in terms of overall Best Acc on different real-world datasets. 
In general, Figure \ref{exp_pic1} shows our proposed Interactive Reinforced Feature Selection (IRFS) method achieves the best overall performance on eight datasets. 
The basic reinforced feature selection method (MARLFS) has better performance than traditional feature selection methods in most cases, because the reinforced feature selection could globally optimize the feature subspace exploration. Moreover, our IRFS shows further improvement compared to MARLFS, which shows it is more effective in feature selection.

\vspace{-3mm}
\subsection{Study of Interactive Reinforced Feature Selection}

We aim to study the impacts of our interactive reinforced feature selection methods on the efficiency of exploration. Our main study subjects include KBest based trainer, Decision Tree based trainer and Hybrid Teaching strategy. 
Accordingly, we consider four different methods: 
(i) \textbf{MARLFS}: The basic reinforced feature selection method, which could be seen as a variant of IRFS without any trainer.
(ii) IRFS with \textbf{KBT} (KBest based Trainer): a variant of interactive reinforced feature selection with only KBT as the trainer. 
(iii) IRFS with \textbf{DTT} (Decision Tree based Trainer): a variant interactive reinforced feature selection with only DTT as the trainer. 
(iv) IRFS with \textbf{HT} (Hybrid Teaching strategy): a variant of interactive reinforced feature selection using Hybrid Teaching strategy with two proposed trainers.

Figure \ref{exp_pic2} shows the comparisons of best accuracy over exploration steps on four datasets. We can observe that both KBT and DTT could reach higher accuracy than MARLFS in few steps (2000 steps), which signifies the trainer's guidance improves the exploration efficiency by speeding up the process of finding optimal subsets. Also, IRFS with HT could find better subsets in short-term compared to MARLFS, KBT and DTT. A potential interpretation is Hybrid Teaching strategy integrates experience from two trainers and takes advantage of broader range of knowledge, leading to the further improvement.

\vspace{-3mm}
\subsection{Study of Varaint Methods}

In this section, we aim to study the impacts of our proposed state representation methods as well as our personalized reward schemes. For this aim, we consider the following five variant methods:
(i) IRFS with \textbf{SRDT1} (State Representation with Decision Tree method 1): a variant of IRFS with HT which uses the state representation method 1 in Section \ref{sec_state_representation}.
(ii) IRFS with \textbf{SRDT2} (State Representation with Decision Tree method 2): a variant of IRFS with HT which uses the state representation method 2 in Section \ref{sec_state_representation}.
(iii) IRFS with \textbf{PRS1} (Personalized Reward Scheme 2): a variant of IRFS with HT which measures reward with decision tree structure feedback, detailed in \ref{section_reward-scheme1}.
(iv) IRFS with \textbf{PRS2} (Personalized Reward Scheme 1): a variant of IRFS with HT which measures reward with historical action records, detailed in \ref{section_reward_scheme2}.
(v) IRFS with \textbf{SRTD+PRS}: a variant of IRFS with HT which uses decision tree structure feedback for state representation and personalized reward scheme.

\begin{figure*}[htbp]
\centering
\subfigure[PRD Dataset]{
\includegraphics[width=4.1cm]{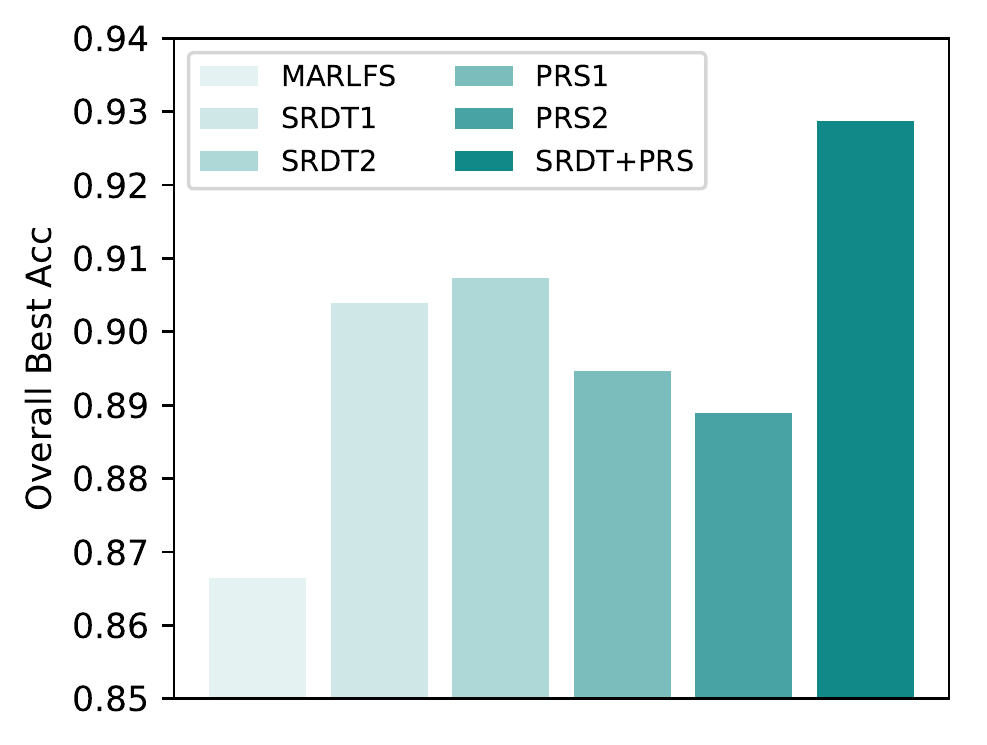}
}
\hspace{-0mm}
\subfigure[FC Dataset]{
\includegraphics[width=4.1cm]{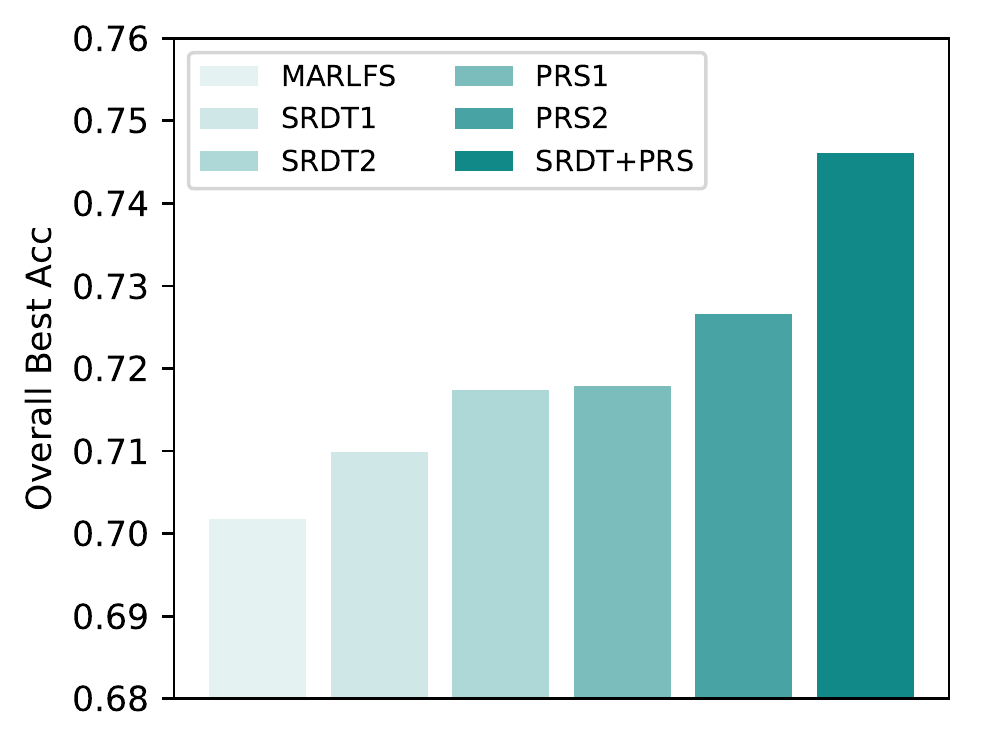}
}
\hspace{-0mm}
\subfigure[Spam Dataset]{
\includegraphics[width=4.1cm]{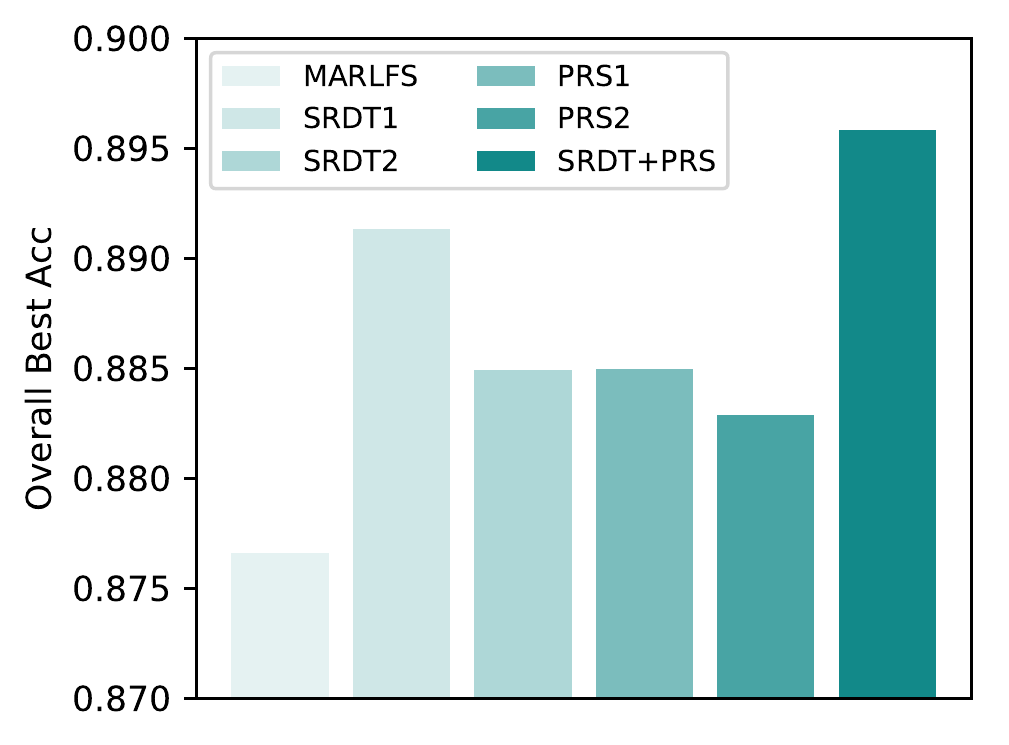}
}
\hspace{-0mm}
\subfigure[ICB Dataset]{
\includegraphics[width=4.1cm]{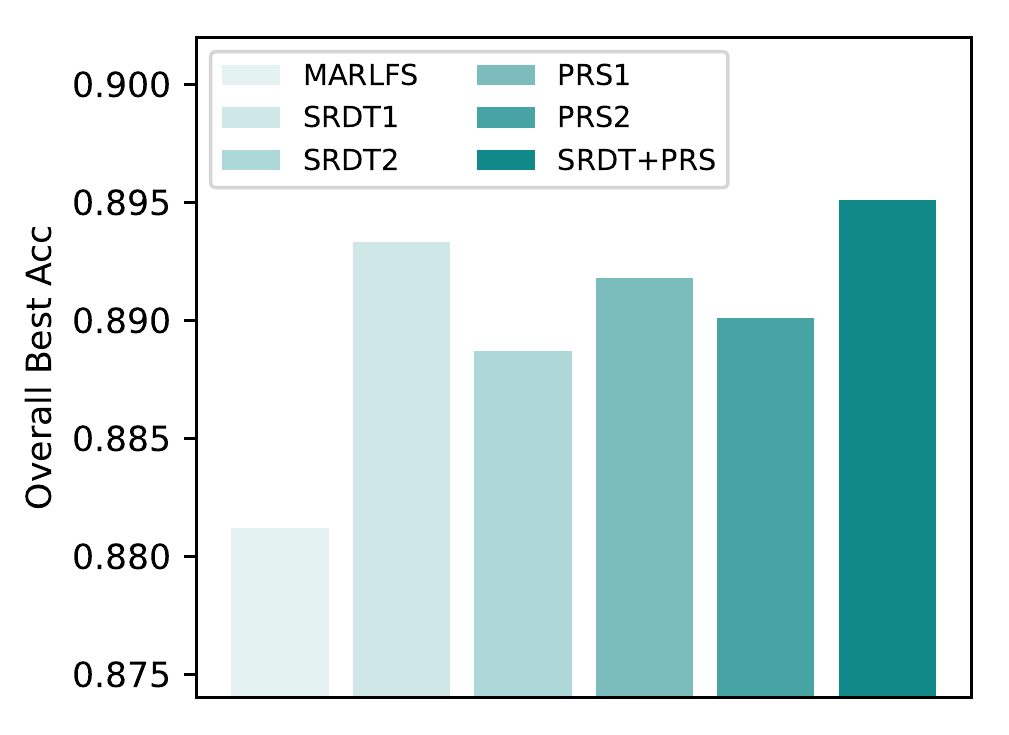}
}
\hspace{-0mm}
\subfigure[Nomao Dataset]{
\includegraphics[width=4.1cm]{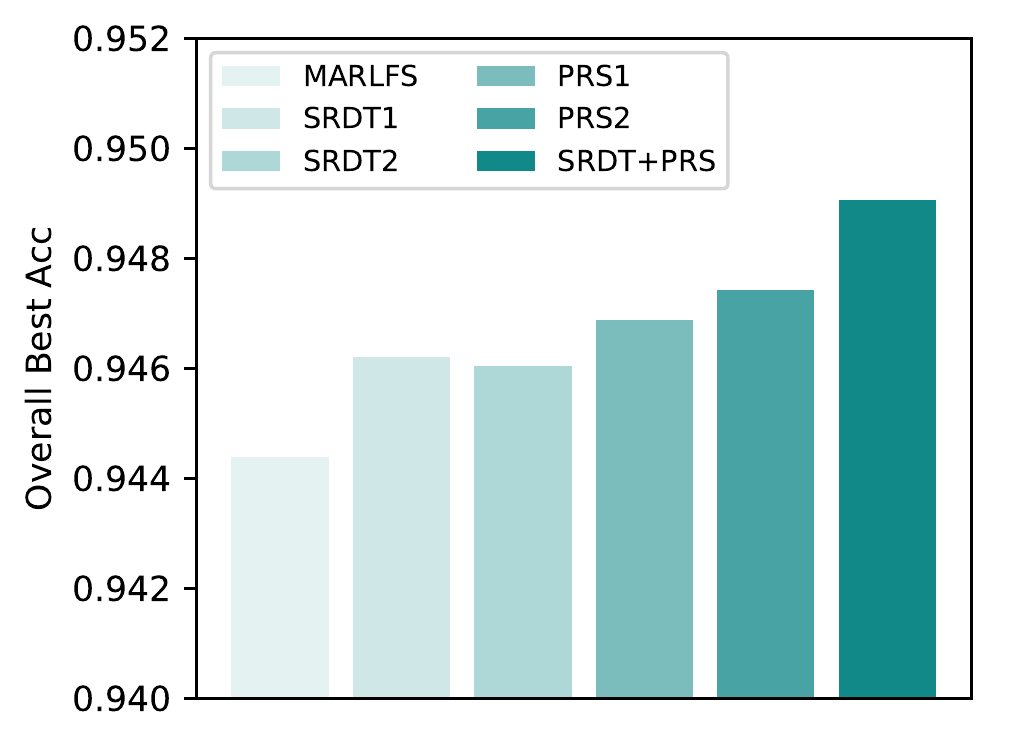}
}
\hspace{-0mm}
\subfigure[Musk Dataset]{
\includegraphics[width=4.1cm]{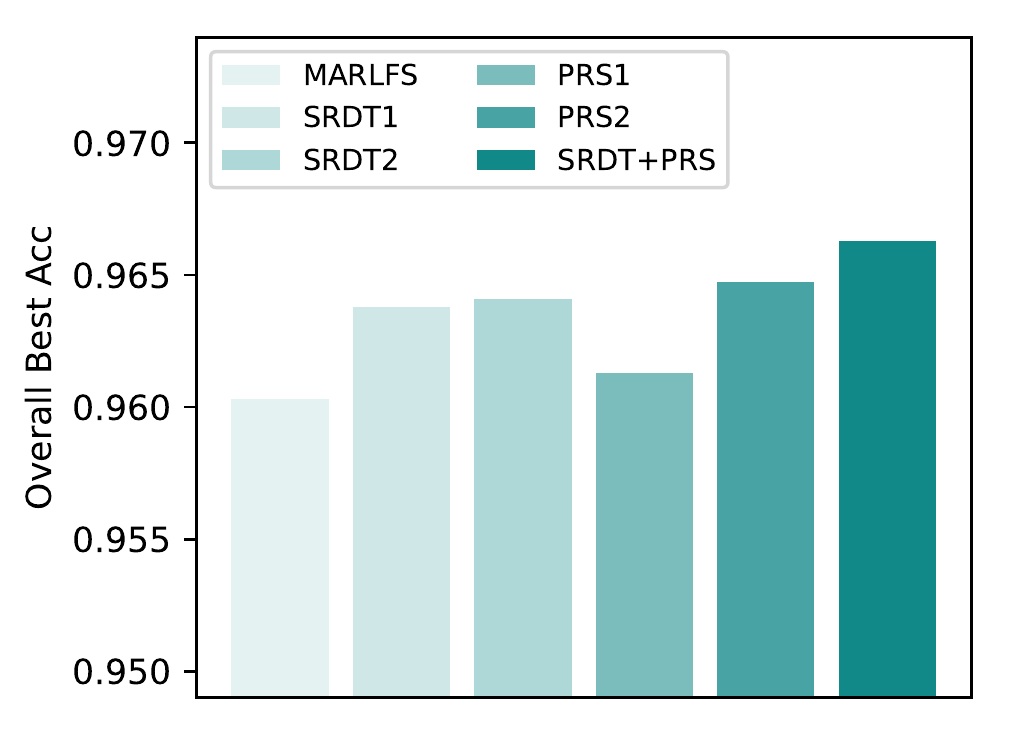}
}
\hspace{-0mm}
\subfigure[ESR Dataset]{
\includegraphics[width=4.1cm]{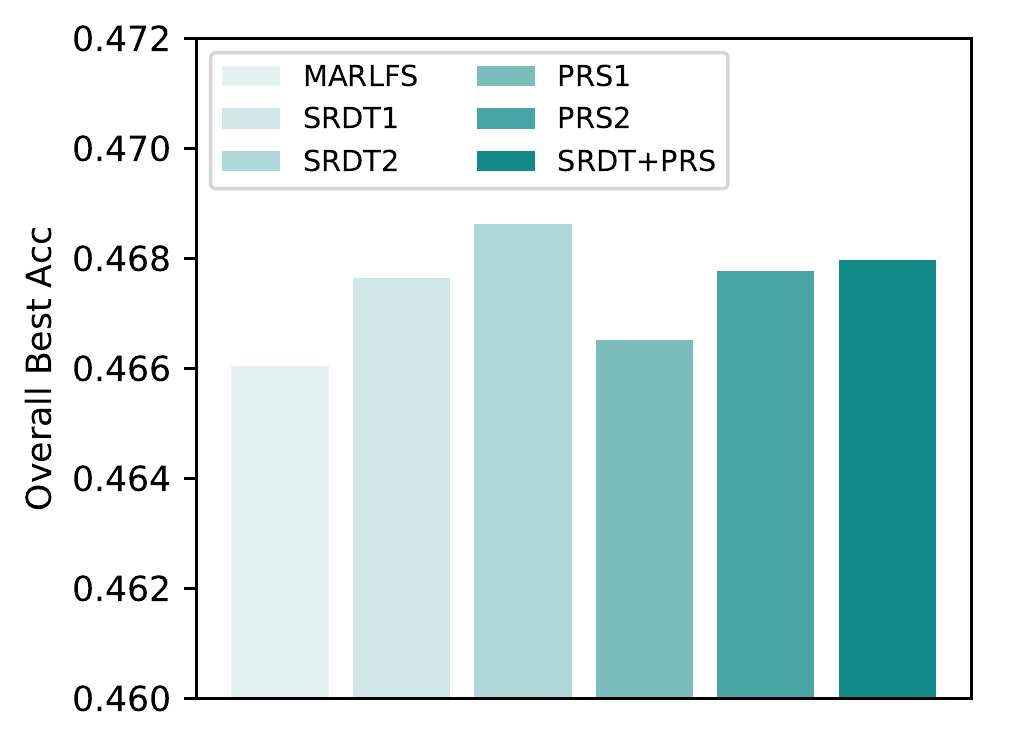}
}
\hspace{-0mm}
\subfigure[QSAR Dataset]{
\includegraphics[width=4.1cm]{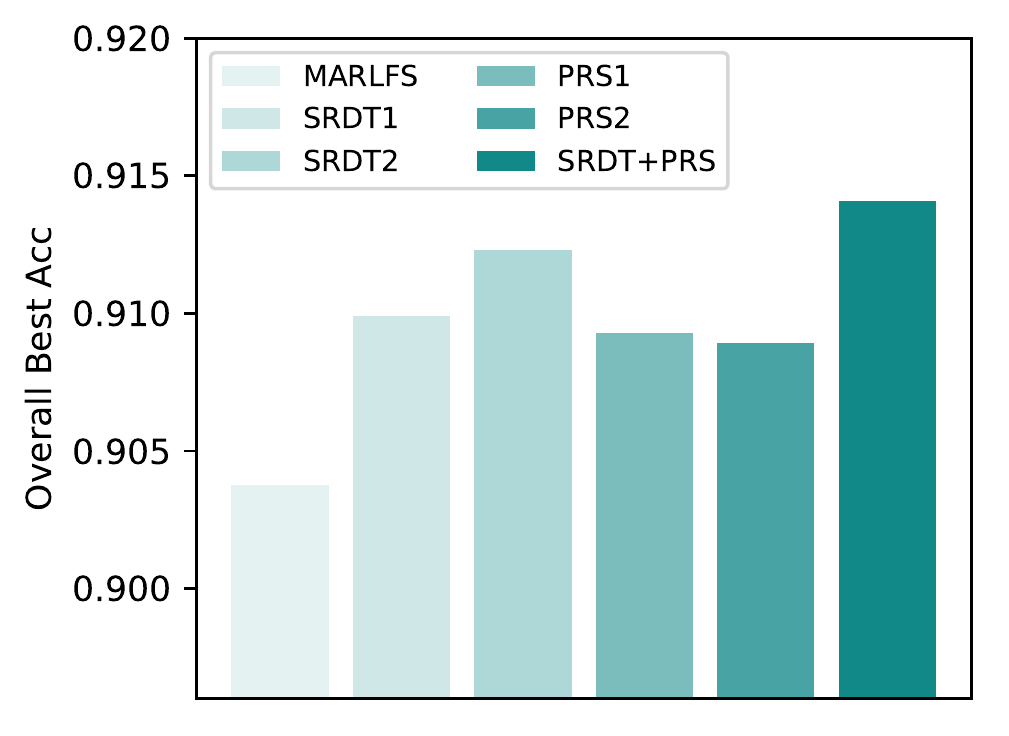}
}
\vspace{-2mm}
\caption{Ave Acc comparison of variant methods. }
\label{exp_pic4}
\vspace{-6mm}
\end{figure*}

We present the performance comparison of these variant methods in terms of Best Acc and Ave Acc. Figure \ref{exp_pic3} shows the Best Acc comparison of different methods. Though the performance differs from methods to methods, in all cases we observe an improvement with respect to the MARLFS baseline. The improved performance reveals that with the help of our proposed state representation methods and reward schemes, IRFS could always find better feature subsets. Moreover, the combination of SRTD and PRS yields better results, which signifies further improved effectiveness in feature selection. We also evaluate our methods in terms of Ave Acc on different datasets. Figure \ref{exp_pic4} shows the result of average performance, where we observe all the proposed methods exceed the basic MARLFS. The improvements of  SRDT and PRS signify precise measurement of state and reward could help for higher average exploration quality, which is important to effective exploration process.

\begin{figure*}[htbp]
\centering
\subfigure[PRD Dataset]{
\includegraphics[width=4.0cm]{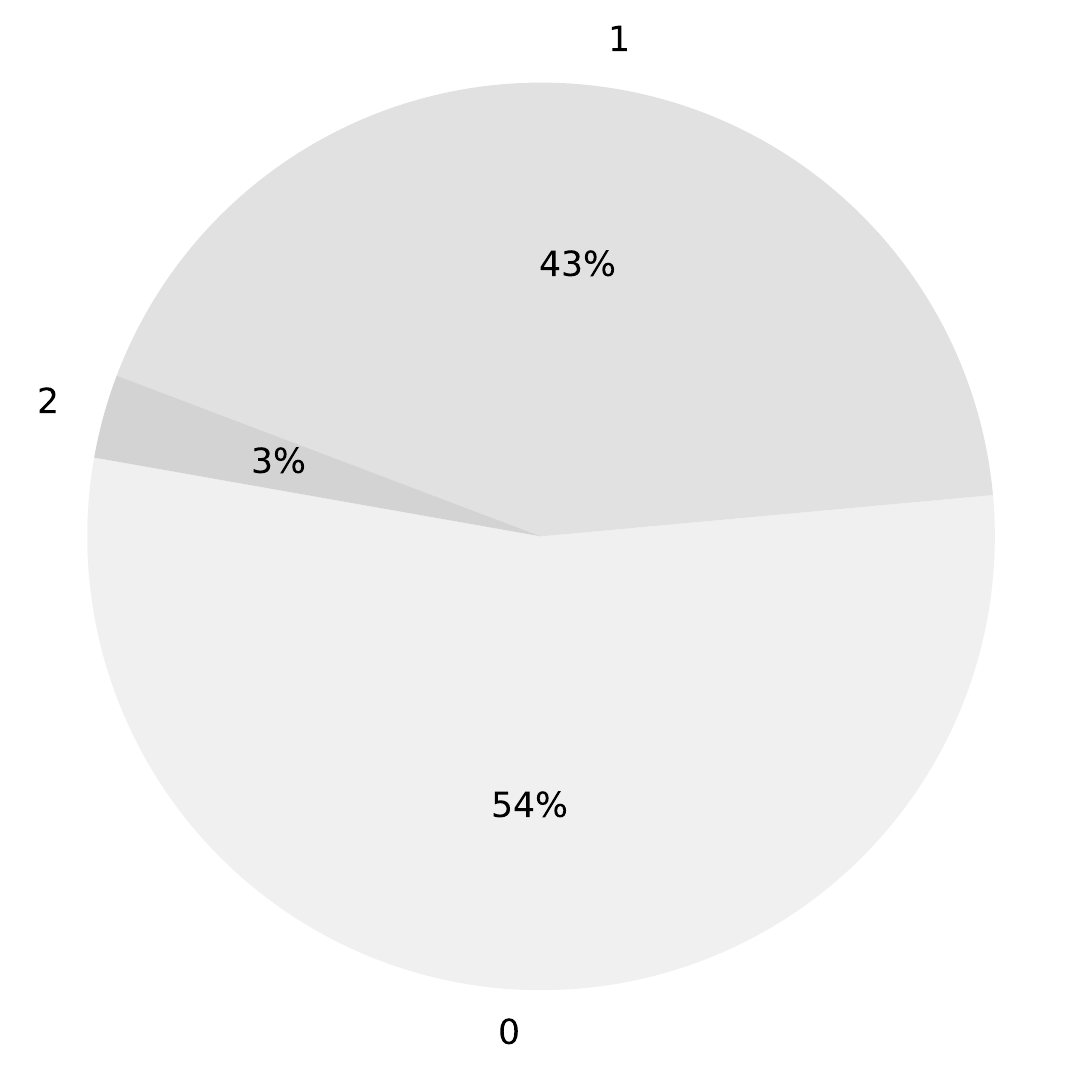}
}
\hspace{-0mm}
\subfigure[FC Dataset]{
\includegraphics[width=4.1cm]{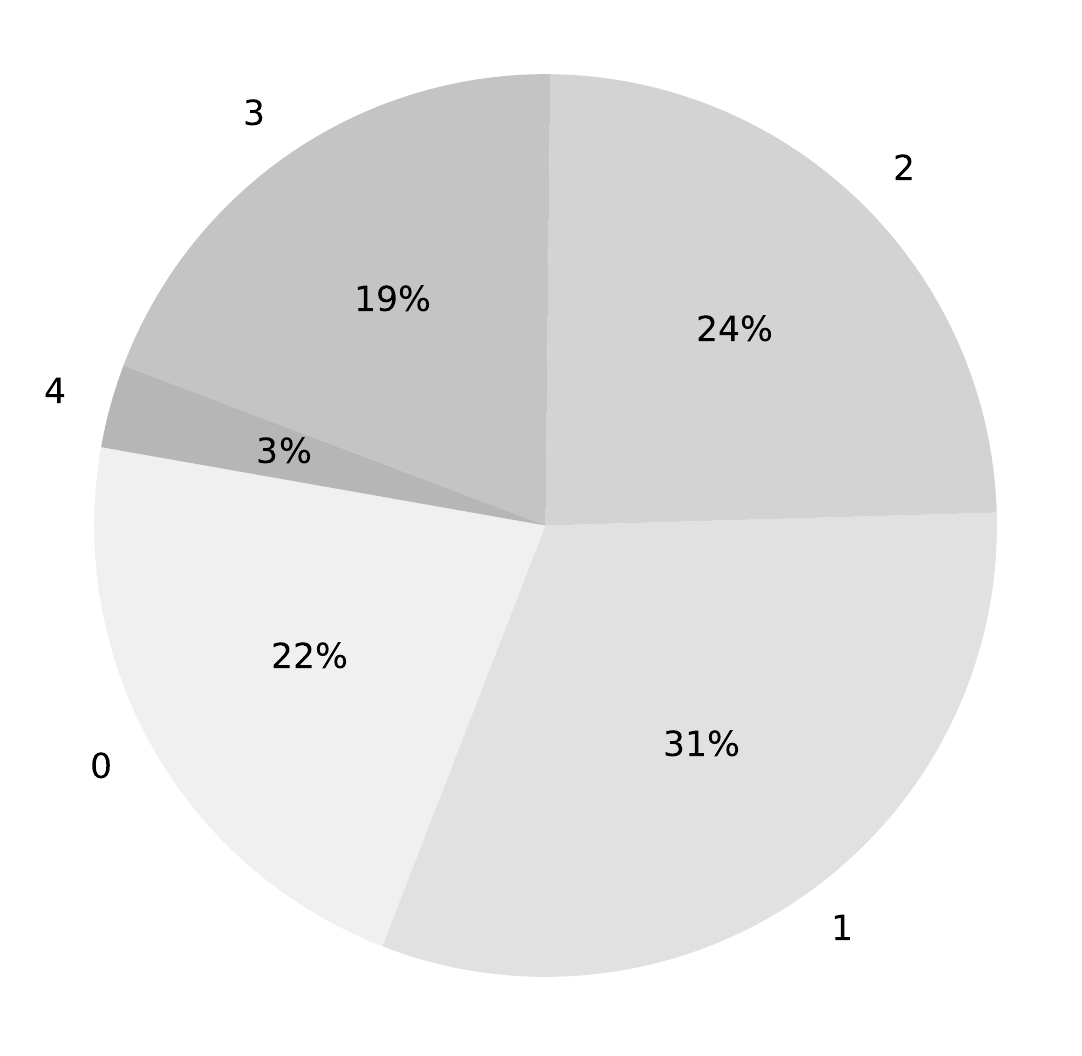}
}
\hspace{-0mm}
\subfigure[Spam Dataset]{
\includegraphics[width=4.1cm]{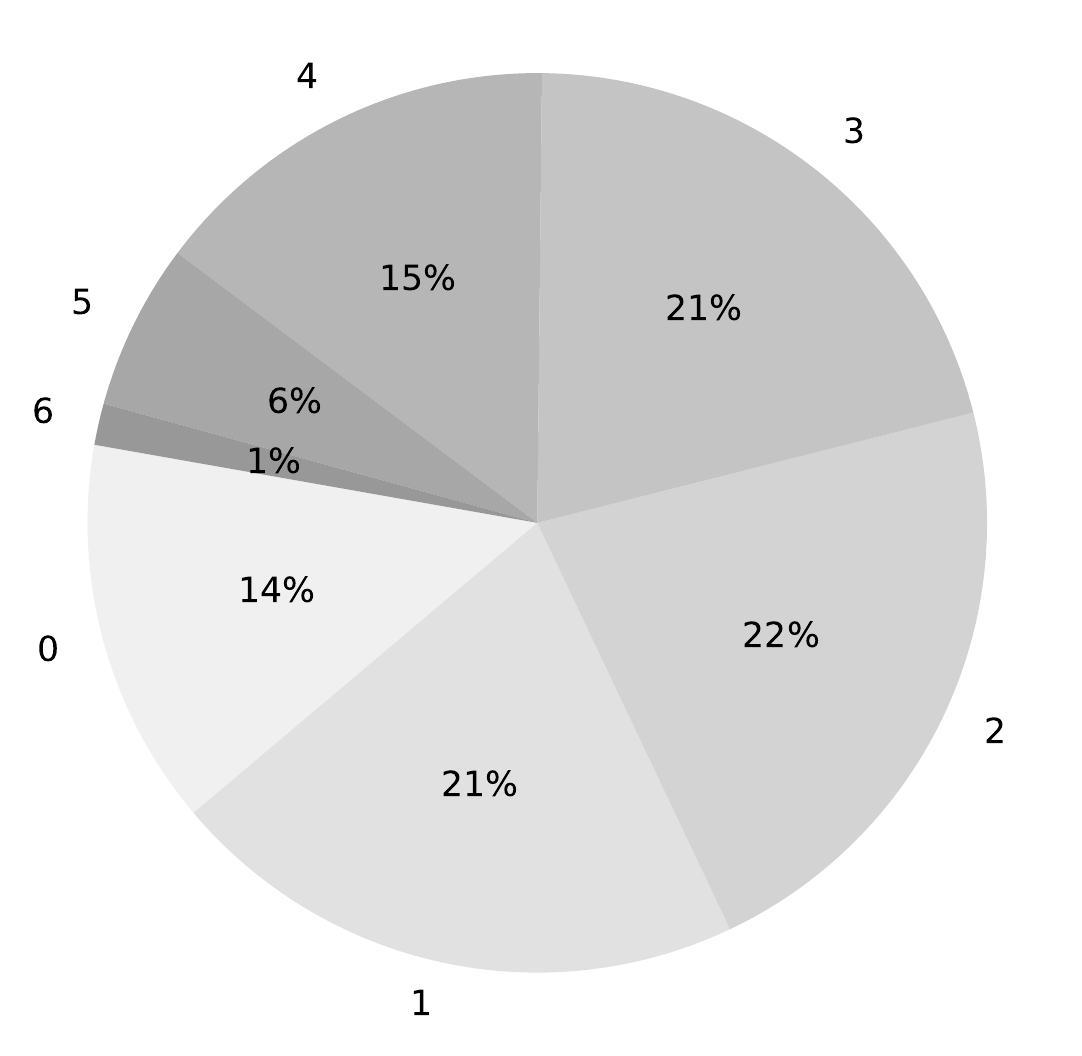}
}
\hspace{-0mm}
\subfigure[ICB Dataset]{
\includegraphics[width=4.1cm]{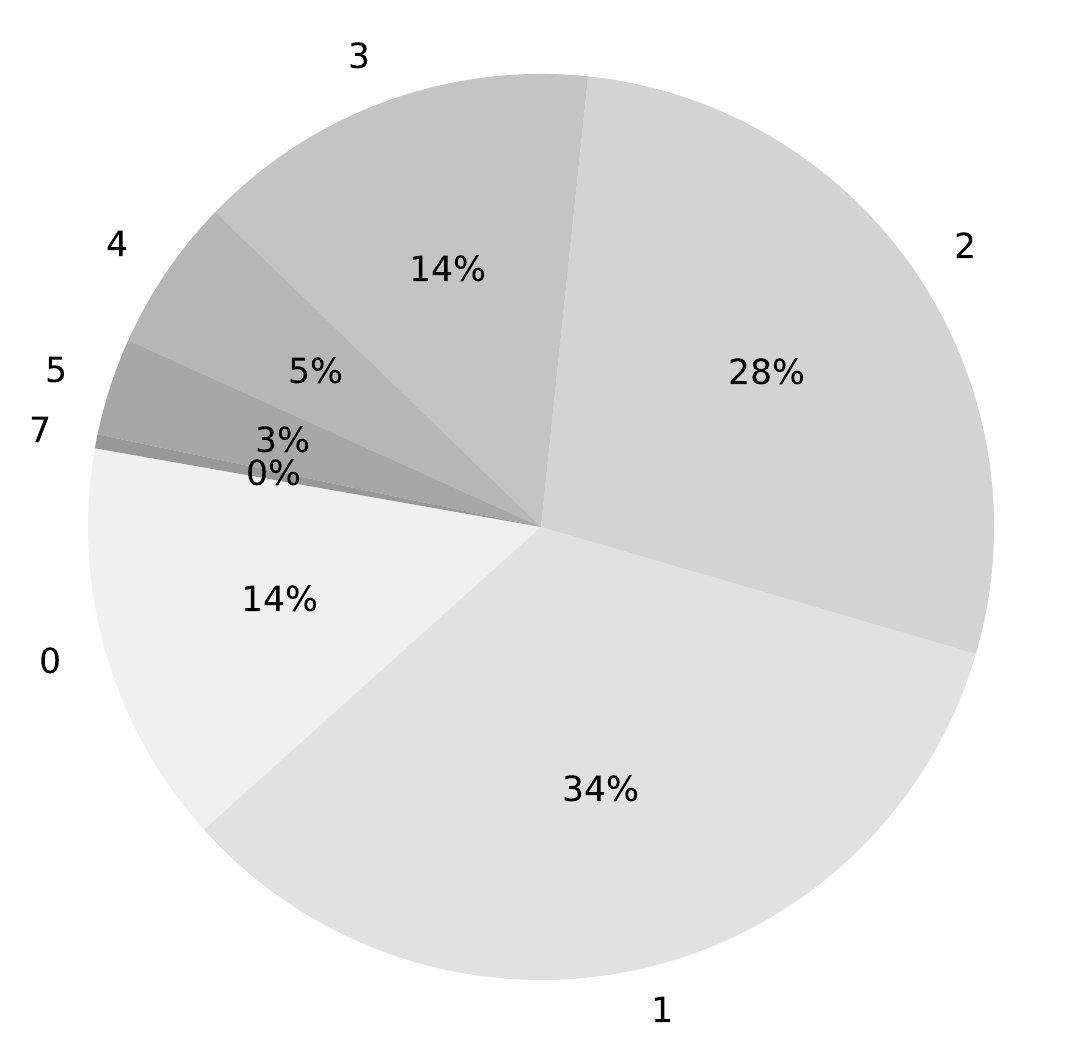}
}
\hspace{-0mm}
\subfigure[Nomao Dataset]{
\includegraphics[width=4.1cm]{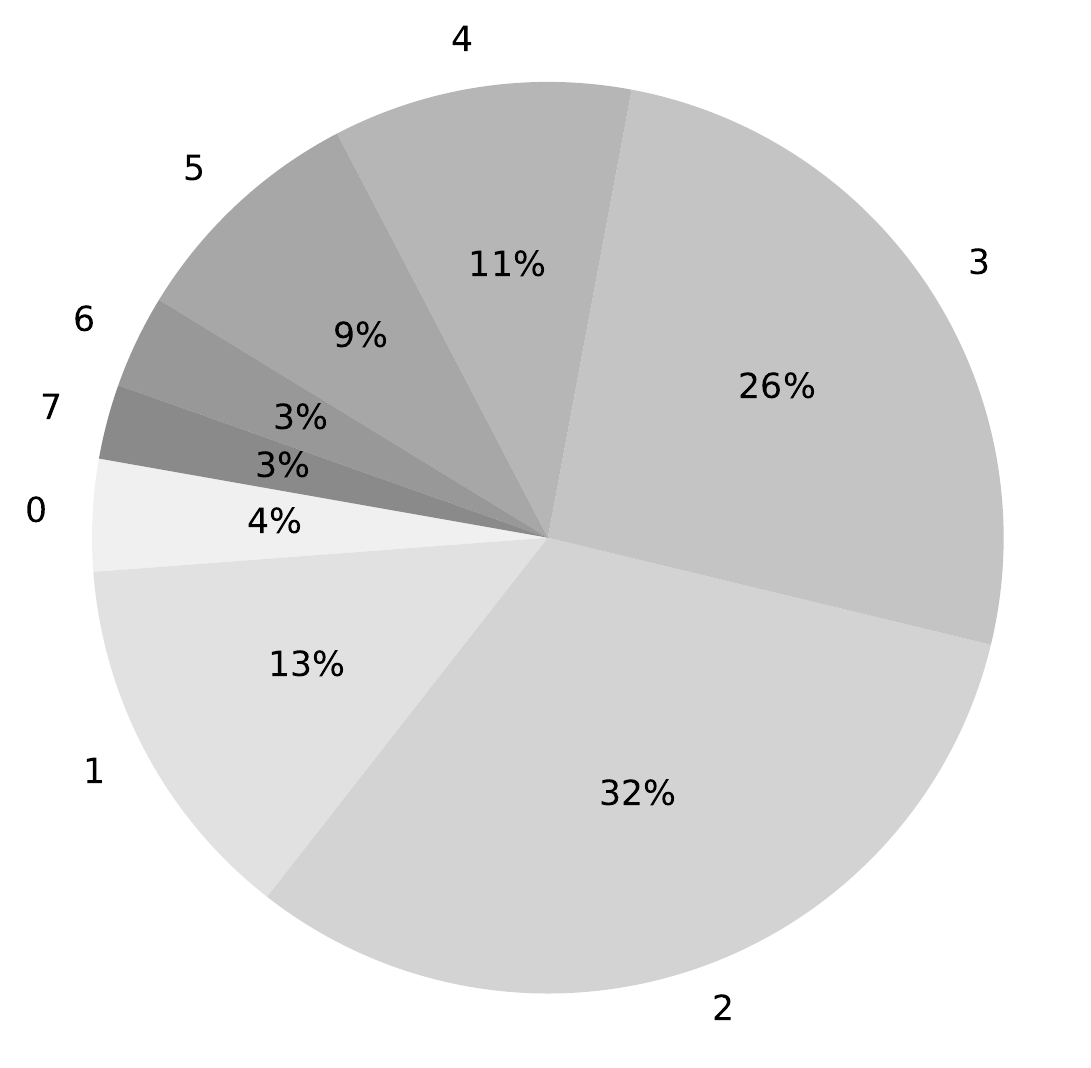}
}
\hspace{-0mm}
\subfigure[Musk Dataset]{
\includegraphics[width=4.1cm]{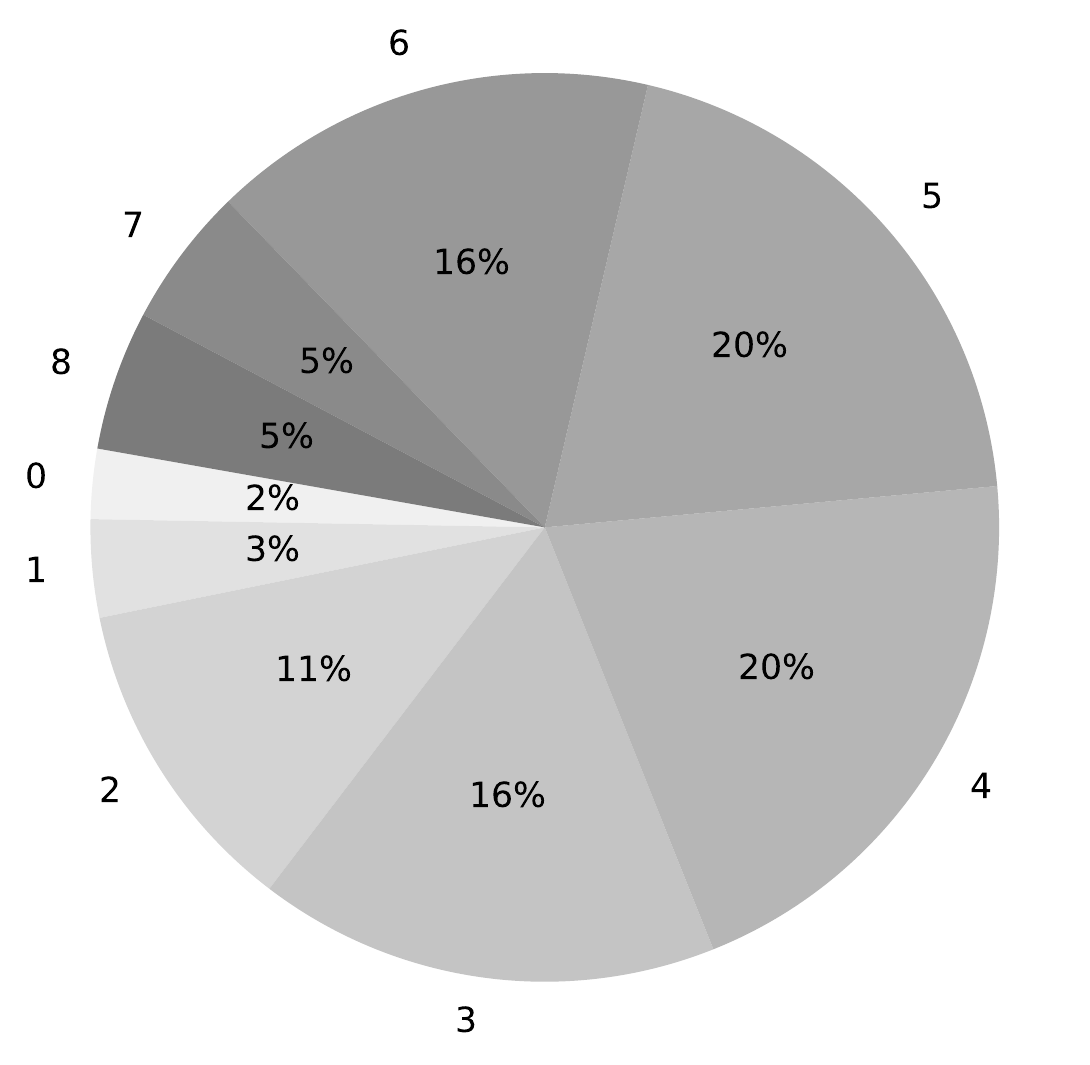}
}
\hspace{-0mm}
\subfigure[ESR Dataset]{
\includegraphics[width=4.1cm]{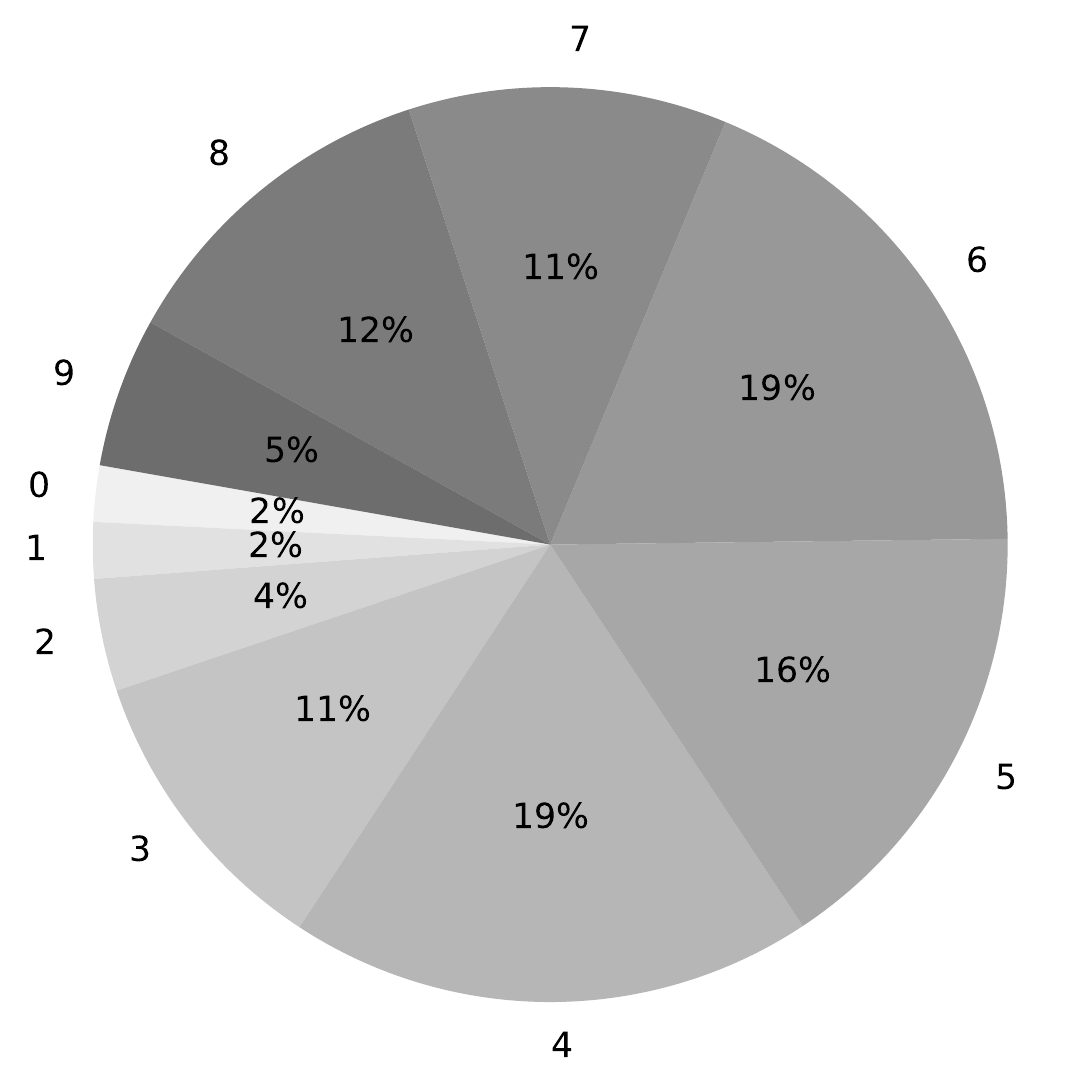}
}
\hspace{-0mm}
\subfigure[QSAR Dataset]{
\includegraphics[width=4.1cm]{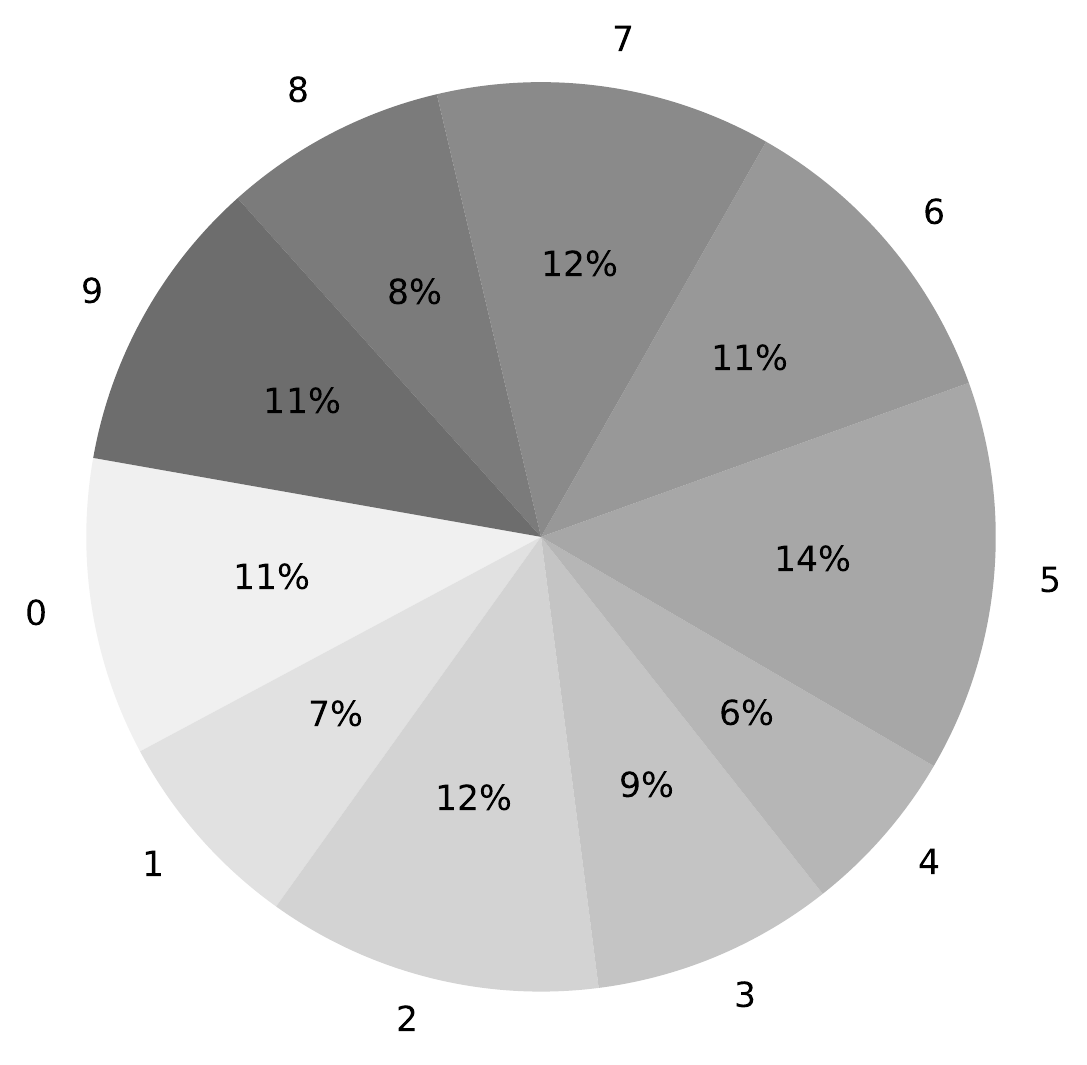}
}
\vspace{-2mm}
\caption{Percentage of action-changed agents in exploration. 0-9 denote the number of action-changed agents.}
\label{exp_pic5}
\vspace{-4mm}
\end{figure*}

\vspace{-1mm}
\subsection{Study of action-changed agents}
In each exploration step, we record the number of hesitant agents that take advice and change their actions from deselection to selection. Figure \ref{exp_pic5} shows the percentage of action-changed agents in the exploration, where 0 to 9 denotes the number of action-changed agents. For example, in Figure 13(e), in 32\% of the exploration steps, there are 2 agents that change their actions. We observe in most steps the number of agents that change actions is non-zero, which reveals the improved performance is due to these advised actions that actually work for better exploration. Another observation is ESR and QSAR Dataset is darker than other datasets, demonstrating there are more action-changed agents. This is because the input features of datasets are more than other datasets, and our trainers would give more advice to agents.

\vspace{-0.3cm}
\section{Related Work}
\vspace{-0.1cm}
Traditional feature selection can be grouped into three kinds of methods: filter methods, wrapper methods, and embedded methods. (1) Filter methods calculate relevance scores, rank features and then select top-ranking ones. Two classical methods are univariate feature selection \cite{forman2003extensive,yang1997comparative} and correlation based feature selection \cite{yu2003feature}. (2) Wrapper methods make use of predictors, considering the prediction performance as objective function \cite{guo2007semantic}. The representative wrapper methods are branch and bound algorithms \cite{kohavi1997wrappers,narendra1977branch}. (3) Embedded methods take more advantages of predictors, incorporation feature selection as part of predictors. The representative methods is LASSO \cite{tibshirani1996regression} and decision tree \cite{sugumaran2007feature}.  
Filter methods are efficient because of the { low computational complexity}, but do not consider the correlation among features. Wrapper methods search on the whole feature subspace, so it may have better performance. But they are computationally expensive because the feature subspace increases exponentially with the increase of number of features. Embedded methods could achieve much  better performance. However, their compatibility with many other predictors is limited.

Reinforcement learning \cite{sutton2018reinforcement} has been applied into different domains, such as robotics \cite{lin1991programming}, urban computing \cite{wang2020incremental}, and feature selection.
Reinforced feature selection applies reinforcement learning for feature selection \cite{fard2013using,kroon2009automatic,bouneffouf2017context}. Some existing studies create a single agent to make decisions \cite{ fard2013using,kroon2009automatic,zhao2020simplifying}.  
However, 
this agent has to determine to select or deselect of all $N$ features, whose action space is $2^N$ and is too large. 
Another kind of existing studies create multi-agents to make decisions and every agent determines the selection of its corresponding feature \cite{liu2019automating}. 

Reinforcement Learning \cite{sutton2018reinforcement} is a good method to address optimal decision-making problem by developing action strategies, whose goal is to maximize the collected reward. In reinforcement learning, the next action is from the highest state-action pair. An intuitive idea to speed up the learning process is to include external advice in the apprenticeship \cite{cruz2018improving}. Actually, interaction mechanism has been studied and applied \cite{liu2018modeling}; in Interactive Reinforcement Learning (IRL), an action is interactively encouraged by a trainer with prior knowledge \cite{knox2013teaching,thomaz2006reinforcement}. Using a trainer to directly advise on future actions is known as policy shaping \cite{amir2016interactive,cederborg2015policy}. For advice, they could come from humans and robots in early studies \cite{lin1991programming}. Other researchers use an artificial trainer-agent which was previously trained to provide advice \cite{taylor2014reinforcement}.

\vspace{-0.3cm}
\section{Conclusion}
\vspace{-0.1cm}
In this paper, we studied the problem of balancing the effectiveness and efficiency of automated feature selection. We first formulate feature selection problem into a diversity-aware interactive reinforcement learning framework, where we develop  a  joint and interactive  architecture  to  unify  both  interaction  between  agents  and  external  trainers, and  interaction  between downstream  task  and  reinforcement learning. 
To measure reward better, we design two new reward schemes that personalize the reward assignment. Finally, we present extensive experiments which illustrate the improved performances.

\vspace{-2mm}
\bibliographystyle{IEEEtran}
\bibliography{ref} 

\end{document}